\documentclass[a4paper,fleqn]{cas-dc}

\usepackage[numbers]{natbib}
\usepackage{amsthm}
\usepackage{graphicx}
\theoremstyle{definition}

\def\tsc#1{\csdef{#1}{\textsc{\lowercase{#1}}\xspace}}
\tsc{WGM}
\tsc{QE}
\tsc{EP}
\tsc{PMS}
\tsc{BEC}
\tsc{DE}

\begin{document}
\let\WriteBookmarks\relax
\def\floatpagepagefraction{1}
\def\textpagefraction{.001}
\shorttitle{Dimensions of Commonsense Knowledge}
\shortauthors{Ilievski et~al.}

\title [mode = title]{Dimensions of Commonsense Knowledge}
% \tnotemark[1,2]

% \tnotetext[1]{This document is the results of the research
%   project funded by the National Science Foundation.}

% \tnotetext[2]{The second title footnote which is a longer text matter
%   to fill through the whole text width and overflow into
%   another line in the footnotes area of the first page.}

\author[1]{Filip Ilievski}
\cormark[1]
\ead{ilievski@isi.edu}
%\ead[url]{www.cvr.cc, cvr@sayahna.org}

%\credit{Conceptualization of this study, Methodology, Software}

\address[1]{Information Sciences Institute, University of Southern California, Marina del Rey, CA, USA}

\author[2]{Alessandro Oltramari}
\ead{Alessandro.Oltramari@us.bosch.com}
%\ead[URL]{www.sayahna.org}

%\credit{Data curation, Writing - Original draft preparation}

\address[2]{Intelligent Internet of Things, Bosch Research and Technology Center, Pittsburgh, PA, USA}

\author%
[3]
{Kaixin Ma}
\ead{kaixinm@andrew.cmu.edu}

\address[3]{Language Technologies Institute, Carnegie Mellon University, Pittsburgh, PA, USA}

\author%
[1]
{Bin Zhang}
\ead{binzhang@isi.edu}

\author%
[4]
{Deborah L. McGuinness}
\ead{dlm@cs.rpi.edu}
%orcid - https://orcid.org/0000-0001-7037-4567

\address[4]{Tetherless World Constellation, Rensselaer Polytechnic Institute, Troy, NY, USA}

\author%
[1]
{Pedro Szekely}
\ead{pszekely@isi.edu}

\cortext[cor1]{Corresponding author}

% \nonumnote{This note has no numbers. In this work we demonstrate $a_b$
%   the formation Y\_1 of a new type of polariton on the interface
%   between a cuprous oxide slab and a polystyrene micro-sphere placed
%   on the slab.
%   }

\begin{abstract}
Commonsense knowledge is essential for many AI applications, including those in natural language processing, visual processing, and planning. Consequently, many sources that include commonsense knowledge have been designed and constructed over the past decades. Recently, the focus has been on large text-based sources, which facilitate easier integration with neural (language) models and application to textual tasks, typically at the expense of the semantics of the sources and their harmonization. Efforts to consolidate commonsense knowledge have yielded partial success, with no clear path towards a comprehensive solution. We aim to organize these sources around a common set of dimensions of commonsense knowledge. We survey a wide range of popular commonsense sources with a special focus on their relations. We consolidate these relations into 13 knowledge dimensions. This consolidation allows us to unify the separate sources and to compute indications of their coverage, overlap, and gaps with respect to the knowledge dimensions. Moreover, we analyze the impact of each dimension on downstream reasoning tasks that require commonsense knowledge, observing that the temporal and desire/goal dimensions are very beneficial for reasoning on current downstream tasks, while distinctness and lexical knowledge have little impact. These results reveal preferences for some dimensions in current evaluation, and potential neglect of others.

\end{abstract}

% \begin{graphicalabstract}
% \includegraphics{figs/grabs.pdf}
% \end{graphicalabstract}

% \begin{highlights}
% \item Research highlights item 1
% \item Research highlights item 2
% \item Research highlights item 3
% \end{highlights}

\begin{keywords}
commonsense knowledge \sep semantics \sep knowledge graphs \sep reasoning
\end{keywords}
%todo - where are these keywords coming from?  these need to be updated

\maketitle

% Plan:
% \begin{enumerate}
%     % \item PS to add object examples (on food) for each dimension
%     % \item AO to take over
%     % \item FI to fill 2.2.1
%     % \item AO to work on the theories of knowledge (3.1)
%     % \item DLM to start describing the dimensions in sec 4
%     % \item think of the story - dimensions and sources - which comes first
%     % \item PS finish the table (constraint: everything comes from a source)
%     % \item FI to attempt to work on the story of section 3 - what to put in 3.3
%     \item AO to make section 5.1 coherent, add references; also see if .2-.3 are needed and what is their role
%     \item DLM to resolve some todos and adapt comments in section 3
%     % \item FI to make a pass on sections 1-4
%     % \item PS to complete his examples and insert the table
%     \item FI to work on section 3 and insert the table from PS %https://docs.google.com/spreadsheets/d/1oc5WAhIojnWnZNuPOpcgmTDVZwFZguacs\_JIliePRGU/edit\#gid=0
% \end{enumerate}

\section{Introduction}

\textbf{Commonsense knowledge} is information that humans
typically have that helps them make sense of everyday situations. 
As such, this knowledge can generally be assumed to be possessed by most people, and, according to the Gricean maxims~\cite{grice1975logic}, it is typically omitted in (written or oral) communication. 
The fact that commonsense knowledge is often implicit presents a challenge for automated natural language processing (NLP) and question answering (QA) approaches as the extraction and learning algorithms cannot count on the commonsense knowledge being available directly in text.

Due to its prominence and implicit nature, capturing commonsense knowledge holds a promise to benefit various AI applications, including those in NLP, computer vision, and planning. For instance, commonsense knowledge can be used to fill gaps and explain the predictions of a (neural) model~\cite{lin2019kagnet}, understand agent goals and causality in stories~\cite{williams2017understanding}, or enhance robot navigation and manipulation~\cite{yang2018visual}. Its benefits have already been shown in downstream applications, such as software requirements localization~\cite{Onyeka2020UsingCK} and aligning local laws with public reactions~\cite{Puri2018PragmaticsAS}.

Consequently, acquiring and representing commonsense knowledge in a machine-readable form, as well as reasoning with it, has been a major pursuit of AI since its early days~\cite{mccarthy1960programs}. 
This has manifested in 
the design, construction, and curation of a rich palette of
resources that include commonsense information (potentially along with other content)
like Cyc~\cite{lenat1995cyc}, ATOMIC~\cite{sap2019atomic}, WebChild~\cite{tandon2017webchild}, ConceptNet~\cite{speer2017conceptnet}, WordNet~\cite{miller1998wordnet}, FrameNet~\cite{baker1998berkeley}, and 
Visual Genome \cite{krishna2017visual}. Some of these, such as ConceptNet and Cyc, have been deliberately created to capture
information that would be useful for common sense-related reasoning tasks, while others, like WordNet or Visual Genome, were intended to support other tasks such as word sense disambiguation or image object recognition. As reported in~\cite{ilievski2020consolidating}, the commonsense sources exhibit great diversity in terms of their representational formats, creation methods, and coverage. While there is an opportunity for this knowledge to be exploited jointly, the inherent diversity makes the consolidation of these sources challenging.

Meanwhile, the last few years have featured an increased focus on benchmarks that can be used to evaluate different aspects of common sense, including social~\cite{sap-etal-2019-social}, physical~\cite{bisk2020piqa}, visual~\cite{zellers2019recognition}, and numeric~\cite{lin2020birds} common sense. Further distinction has been made between discriminative tasks~\cite{sap-etal-2019-social,bisk2020piqa,talmor-etal-2019-commonsenseqa}, where the goal is to pick the single correct answer from a list, and generative tasks, where one has to generate one or multiple correct answers~\cite{lin2020birds,boratko2020protoqa}. These tasks can be tackled by using (the entire or a subset of the) training data~\cite{ma2019towards,lin2019kagnet}, or in a zero-/few-shot evaluation regime~\cite{ma2020knowledgedriven,shwartz2020unsupervised}.

The wealth and diversity of commonsense sources, on one hand, and benchmarks, on the other hand, raises a natural question: what is the role of these knowledge repositories in real-world reasoning techniques that need to incorporate commonsense knowledge? While intuitively such sources of commonsense knowledge can have tremendous value for reasoning in downstream reasoning tasks, the practice shows that their impact on these tasks has been relatively limited, especially in comparison to the contribution of the language models (LMs).\footnote{This phenomenon can be seen on the benchmark leadearboards, which are dominated by `pure' language models, for instance: \url{https://leaderboard.allenai.org/socialiqa/submissions/public} (accessed on January 5th, 2021).}
Knowledge sources tend to have larger contributions when little or no training data is available: for example, one can generate artificial training sets based on several sources, which can be used to pre-train language models and apply them on downstream tasks without using the official training data~\cite{banerjee2020self,ma2020knowledgedriven}.
The impact of knowledge resources so far has been generally restricted to the special cases where the knowledge and the task are known (in advance) to be well-aligned~\cite{ma2020knowledgedriven,ma2019towards}. 
While a variety of sources~\cite{sap2019atomic,miller1998wordnet,speer2017conceptnet} or their combination~\cite{ilievski2020consolidating} have been used to enhance language models for downstream reasoning, little is known about how this alignment between knowledge types and tasks can be dynamically achieved.

Most recent sources have focused on the breadth of knowledge, sometimes at the expense of its semantics~\cite{bhakthavatsalam2020genericskb,mostafazadeh2020glucose}. Text-based representations are particularly attractive, as they facilitate a more direct integration with language models, as well as reasoning on NLP and QA tasks. These sources are often treated as `corpora', where each fact is typically lexicalized (manually or automatically) into a single sentence~\cite{ma2019towards}, which is used to inform or fine-tune a language model. 
Due to the lack of focus on formal representational principles, the sources capture knowledge types which are not trivial to align with other sources, as shown by the sparse mappings available between these sources~\cite{ilievski2020consolidating}. 
Considering the 
lack of a common vocabulary and the lack of alignment of these sources, their limited coverage, and lack of focus on explicit semantics, 
knowledge is typically kept in an impoverished textual form that is easy to capture and combine with language models. The downsides of this practice are: 1) the commonsense knowledge remains difficult to harmonize across sources; 2) without a thorough harmonization or consolidation, it is not clear how to effectively measure coverage, overlap, or gaps; and 3) text-based representations alone may be unable to match the human-like level of sophistication typical for contextual reasoning.

Efforts to consolidate commonsense knowledge across sources~\cite{ilievski2020consolidating,gangemi2016framester,navigli2012babelnet} have managed to bring these sources closer, which has shown impact on commonsense QA tasks~\cite{ma2020knowledgedriven}. In \cite{ilievski2020commonsense}, we provide heuristics for defining the boundaries of commonsense knowledge, in order to extract such a subset from one of the largest available graphs today, Wikidata~\cite{vrandevcic2014wikidata}. Yet, these efforts have limited success, and many consolidation questions are left open. How should one think about commonsense knowledge in a theoretical way? What does it mean to build a consolidated knowledge graph (KG) of resources created largely in a bottom-up fashion? How should the relations be chosen? What is the right level of abstraction for relations and nodes?

\section{Approach}
\label{sec:approach}

We aim to provide insights into such questions, aiming primarily to organize the types of knowledge found in existing sources of commonsense knowledge.

For this purpose, we survey a wide variety of sources of commonsense knowledge, ranging from commonsense KGs through lexical and visual sources, to the recent idea of using language models or corpora as commonsense knowledge bases. We survey their relations and group them into a set of \textbf{dimensions}, each being a cluster of its specific relations, as found in the sources. We then apply these dimensions to transform and unify existing sources, providing an enriched version of the Commonsense Knowledge Graph (CSKG)~\cite{ilievski2020consolidating}. 
The dimensions allow us to perform four novel experiments:
\begin{enumerate}
    \item We assess the coverage of the sources with respect to each dimension, noting that some sources have  wide (but potentially shallow) coverage of dimensions, whereas others have deep but narrow coverage. This supports the need to integrate these complementary sources into a single one.
    \item The consolidation of the dimensions enables us to compare knowledge between the sources and compute metrics of overlap. The results show that there is little knowledge overlap across sources, even after consolidating the relations according to our dimensions. %This motivates future work on node resolution.
    \item The statements that are expressed with a dimension form a cluster of knowledge. We contrast these dimension-based clusters to language model-based clusters, which allows us to understand some similarities and differences in terms of their focus. 
    \item We measure the impact of each dimension on two representative commonsense QA benchmarks. Following~\cite{ma2020knowledgedriven}, we pre-train a language model and apply it on these benchmarks in a zero-shot fashion (without making use of the task training data). The dimensions provide a more direct alignment between commonsense knowledge and the tasks, revealing that some dimensions of knowledge are far more informative for reasoning on these tasks than others.

\end{enumerate}

The contributions of the paper are as follows. 1) We survey a wide variety of existing sources of commonsense knowledge, with an emphasis on their relations. 
We provide a categorization of those sources and include a short overview of their focus and creation methods~\textbf{(Section~\ref{sec:sources})}. 
2) We analyze the entire set of relations and abstract them to a set of 13 commonsense dimensions. Each dimension abstracts over more specific relations, as found in the sources~\textbf{(Section~\ref{sec:dimensions})}. 3) The identified dimensions are applied to consolidate the knowledge in CSKG, which integrates seven of the sources we analyze in this paper. The resulting resource is publicly available~\textbf{(Section \ref{sec:experiments})}.
4) We make use of this dimension-based enrichment of CSKG to
analyze the overlap, coverage, and knowledge gaps of individual knowledge sources in CSKG, motivating their further consolidation into a single resource~\textbf{(Sections \ref{ssec:exp1} - \ref{ssec:exp3})}. 5) We evaluate the impact of different dimensions on 
two popular downstream commonsense reasoning tasks. The results show that certain dimensions, like temporal and motivational knowledge, are very beneficial and well-covered by benchmarks, whereas other dimensions like distinctness and lexical knowledge currently have little impact. These results show that our dimensions can be used to facilitate more precise alignment between the available knowledge and question answering tasks. This evaluation also points to gaps in the current knowledge sources and tasks~\textbf{(Section \ref{ssec:reasoning})}.
6) We reflect on the results of our analysis, and use the resulting observations as a basis to provide a roadmap towards building a more comprehensive resource that may further advance the representation of, and reasoning with, commonsense knowledge. Such a resource would be instrumental in building a general commonsense service in the future~\textbf{(Section \ref{sec:discussion})}.
\section{Sources of Commonsense Knowledge}
\label{sec:sources}

We define a \textbf{digital commonsense knowledge source} as a potentially multi-modal repository from which commonsense knowledge can be extracted.\footnote{For brevity, we omit the word `digital' in the remainder of this paper.} Commonsense knowledge sources come in various forms and cover different types of knowledge. While only a handful of sources have been formally proposed as commonsense sources, many others cover aspects of common sense. Here, we collect a representative set of sources, which have been either proposed as, or considered as, repositories of commonsense knowledge in the past. We categorize them into five groups, and describe the content and creation method of representative sources within each group.
Table~\ref{tab:sources} contains statistics and examples for each source.

\begin{table*}
    \footnotesize
    \centering
       \label{tab:sources}
    \caption{Overview of commonsense knowledge sources. The asterisk (`*') indicates that the source is extended with WordNet knowledge. For FrameNet and MetaNet, we specify their numbers of frame-to-frame relations. WebChild contains a large number of relations, expressed as WordNet synsets, which are aggregated into 4 groups.}
    \scalebox{0.88}{
    \begin{tabular}{c c c c c}
        \bf Category & \bf Source & \bf Relations & \bf Example 1 & \bf Example 2\\ \toprule
        Commonsense KGs & ConceptNet* & 34 & \it food - capable of - go rotten & \it eating - is used for - nourishment\\ 
        & ATOMIC & 9 & \it Person X bakes bread - xEffect - eat food & \it PersonX is eating dinner - xEffect - satisfies hunger\\
        & GLUCOSE & 10 & \multicolumn{2}{c}{\it $Someone_{A}$ makes $Something_A$ (that is food) Causes/Enables
$Someone_A$ eats $Something_A$} \\
        & WebChild & 4 (groups) & \it restaurant food - quality\#n\#1 - expensive & \it eating - type of - consumption\\
        & Quasimodo & 78,636 & \it pressure cooker - cook faster - food & \it herbivore - eat - plants\\
        & SenticNet & 1 & \it cold\_food - polarity - negative & \it eating breakfast - polarity - positive\\
        & HasPartKB & 1 & \it dairy food - has part - vitamin & \it n/a\\
        & Probase & 1 & \it apple - is a - food & \it n/a \\
        & Isacore & 1 & \it snack food - is a - food & \it n/a \\
        \midrule
        Common KGs & Wikidata & 6.7k & \it food - has quality - mouthfeel & \it eating - subclass of - ingestion \\
        & YAGO4 & 116 & \it banana chip - rdf:type - food & \it eating - rdfs:label - feeding\\
        & DOLCE* & 1 & \it n/a & \it n/a \\
        & SUMO* & 1,614 & \it food - hyponym - food\_product & \it process - subsumes - eating \\ \midrule
        Lexical resources & WordNet & 10 & \it food - hyponym - comfort food & \it eating - part-meronym - chewing\\ 
        & Roget & 2 & \it dish - synonym - food & \it eating - synonym - feeding\\
        & FrameNet & 8 (f2f) & \it Cooking\_creation - has frame element - Produced\_food & \it eating - evoke - Ingestion \\
        & MetaNet & 14 (f2f) & \it Food - has role - food\_consumer & \it consuming\_resources - is - eating \\
        & VerbNet & 36 (roles) & \it feed.v.01 - Arg1-PPT - food & \it eating - hasPatient - comestible\\ \midrule
        Visual sources & Visual Genome & 42,374 & \it food - on - plate & \it boy - is eating - treat\\
        & Flickr30k & 1 & \it a food buffet - corefers with - a food counter & \it a eating place - corefers with - their kitchen \\ \midrule
        Corpora \& LMs & GenericsKB & n/a & \it Aardvarks search for food. & \it Animals receive nitrogen by eating plants. \\
        & GPT-2 & n/a & \it Food causes  a person to be hungry and a person to eat. & \it Eating  at home will not lead to weight gain. \\ \bottomrule
        
    \end{tabular}}
\end{table*}

\subsection{Commonsense Knowledge Graphs}

\textbf{ConceptNet}~\cite{speer2017conceptnet} is a multilingual commonsense knowledge graph. Its nodes are primarily lexical and connect to each other with 34 relations. Its data is largely derived from the crowdsourced Open Mind Common Sense corpus~\cite{singh2002open}, and is  complemented with knowledge from other resources, like WordNet.

\textbf{ATOMIC}~\cite{sap2019atomic} is a commonsense knowledge graph that expresses pre- and post-states for events and their participants in a lexical form with nine relations. Its base events are collected from a variety of corpora, while the data for the events is collected by crowdsourcing.

\textbf{GLUCOSE}~\cite{mostafazadeh2020glucose} contains causal knowledge through 10 relations about events, states, motivations, and emotions. The knowledge in GLUCOSE is crowdsourced based on semi-automatic templates, and generalized from individual stories to more abstract rules.

\textbf{WebChild}~\cite{tandon2017webchild} is a commonsense knowledge graph whose nodes and relations are disambiguated as WordNet senses. It captures 20 main relations, grouped in four categories. WebChild has been extracted automatically from Web information, and then is canonicalized in a post-processing step.

\textbf{Quasimodo}~\cite{romero2019commonsense} contains commonsense knowledge about object properties, human behavior, and general concepts, expressed with nearly 80K relations. Its nodes and relations are initially lexical and extracted automatically from search logs and forums, after which a notable subset of them has been clustered into WordNet domains. 

\textbf{SenticNet}~\cite{cambria2020senticnet} is a knowledge base with conceptual and affective knowledge, which is extracted from text and aggregated automatically into higher-level primitives. 

\textbf{HasPartKB}~\cite{bhakthavatsalam2020dogs} is a knowledge graph of hasPart statements, extracted from a corpus with sentences and refined by automatic means.

\textbf{Probase}~\cite{wu2012probase} is a probabilistic taxonomy of IsA statements, which has been automatically extracted from a large corpus.

\textbf{IsaCore}~\cite{cambria2012semantic} is a taxonomy of IsA knowledge selected from ConceptNet and Probase.

\subsection{Common Knowledge Graphs and Ontologies}
\label{ssec:ckgont}

\textbf{Wikidata}~\cite{vrandevcic2014wikidata} is a general-domain knowledge graph, tightly coupled with Wikipedia, that describes notable entities. Its nodes and relations are disambiguated as Qnodes. The content of Wikidata is collaboratively created by humans, as well as other existing sources. Given the vast number of statements in Wikidata and its sizable set of over 7 thousand relations, we consider its \textit{Wikidata-CS} commonsense subset, as extracted by Ilievski et al.~\cite{ilievski2020commonsense}.

\textbf{YAGO}~\cite{pellissier2020yago} is a general-purpose knowledge graph, whose nodes and relations are disambiguated entities. YAGO models its knowledge with 116 relations. The knowledge in YAGO is extracted automatically from Wikipedia, and consolidated with knowledge from other sources, like Schema.org~\cite{guha2016schema}.

\textbf{DOLCE} (Descriptive Ontology for Linguistic and Cognitive Engineering)~\cite{gangemi2002sweetening} is an upper level ontology that captures the ontological categories underlying natural language and human common sense with disambiguated concepts and relations. It has been created manually by experts.

\textbf{SUMO} (Suggested Upper Merged Ontology)~\cite{niles2001towards} is an ontology of upper-level disambiguated concepts and their relations. It has been created manually by experts.

\subsection{Lexical Resources}

\textbf{WordNet}~\cite{miller1998wordnet} is a lexical database of words, their meanings, and taxonomical organization, in over 200 languages. WordNet has nearly a dozen relations. It has been created manually by experts.

\textbf{Roget}~\cite{roget2020roget} is a manually-created thesaurus that contains synonyms and antonyms for English words. 

\textbf{FrameNet}~\cite{baker1998berkeley} is a lexical resource that formalizes the frame semantics theory: meanings are mostly understood within a frame of an event and its participants that fulfill roles in that frame. FrameNet has 8 frame-to-frame relations. FrameNet was created manually by experts.

\textbf{MetaNet}~\cite{dodge2015metanet} is a repository of conceptual frames, as well as their relations which often express metaphors. There are 14 frame-to-frame relations in MetaNet. It has been created manually.

\textbf{VerbNet}~\cite{schuler2005verbnet} is a resource that describes syntactic and semantic patterns of verbs, and organizes them into verb classes. There are 36 relations about roles in VerbNet. VerbNet has been created manually by experts.

\subsection{Visual Commonsense Sources}

\textbf{Visual Genome}~\cite{krishna2017visual} contains annotations of concepts and their relations in a collection of images. VisualGenome has over 40K relations. The image descriptions are manually written by crowd workers, while their concepts are mapped automatically to WordNet senses and revised by crowd workers.

\textbf{Flickr30k}~\cite{plummer2016flickr30k} annotates objects in 30k images by multiple workers. The expressions used by different annotators are clustered automatically into groups of coreferential expressions by van Miltenburg~\cite{miltenburg2016stereotyping}.

\subsection{Corpora and Language Models}

\textbf{GenericsKB}~\cite{bhakthavatsalam2020genericskb} contains self-contained generic facts represented as naturally occurring sentences. The sentences have been extracted from three existing corpora, filtered by handwritten rules, and scored with a BERT-based classifier.

\textbf{Language models}, like RoBERTa~\cite{liu2019roberta} and GPT-2~\cite{radford2019language}, can be used as KGs\cite{petroni2019language} to complement explicitly stated information, e.g., as a link prediction system like COMET~\cite{bosselut2019comet} or through self-talk~\cite{shwartz2020unsupervised}.

\subsection{Observations}

As apparent from the descriptions in this section, the existing commonsense sources are based on a wide range of representation principles and have been created using different construction methods. Through the example scenarios of food and eating (Table~\ref{tab:sources}), we show that they have notable overlap in terms of their covered (typically well-known) concepts. At the same time, the types of knowledge covered differ across sources: some sources provide truisms, such as feeding is done with food, while others speculate on usual properties of food, such as its capability to spoil (or ``go rotten'') or often be placed on a plate. Furthermore, we observe that same or similar relations tend to have different names across sources (compare \textit{type of} to \textit{subclass of} or \textit{is}; or \textit{has quality} in Wikidata to \textit{cook faster} in Quasimodo). 

These distinctions make the integration of these sources, and the understanding of their coverage and gaps, very challenging. In order to integrate the knowledge in these sources, we next propose a consolidation of their relations into a common set of dimensions.

\begin{table*}[!t]
	\centering
	{
	\caption{Knowledge dimensions with their corresponding relations. Relations marked with `*' describe knowledge in multiple dimensions. Relations with $\neg$ express negated statements. Sources in the column \textit{Other}: FN = FrameNet, WN = WordNet, RG = Roget, HP = HasPartKB, PB = Probase, IC = IsaCore, SN = SenticNet.}
	\label{tab:dims}
    \scalebox{0.93}{
	\begin{tabular} {c | c c c c c c}
		\toprule
		\bf Dimension  & \bf ATOMIC & \bf ConceptNet & \bf WebChild & \bf Other & \bf Wikidata \\	
		\midrule
                           & & FormOf \\
        \bf lexical & & DerivedFrom & & lexical\_unit (FN) & label &  \\
                           & & EtymologicallyDerivedFrom & & lemma (WN)				 \\ \hline
                    & & Synonym & & reframing\_mapping (FN) \\
       \bf similarity   & & SimilarTo & hassimilar & metaphor (FN) & \\ 
                    & & DefinedAs &   & Synonym (RG) & said to be the same as	 \\
                    & & & & synonym (WN) & \\ 
                    \hline
         & & Antonym & & Antonym (RG) & different from \\
            \bf distinctness         & & DistinctFrom & & antonym (WN)  & opposite of \\ 
            & & & & excludes (FN) \\ \hline

         & 	& IsA & & perspective\_on (FN) & subClassOf  \\
            \bf taxonomic     &  & InstanceOf &	hasHypernymy & inheritance (FN) & instanceOf  \\
            & & MannerOf & & hypernym (WN) & description \\
            & & & & isA (PB, IC) \\
            \hline
                 & & PartOf & & HasPart (HP) & &  \\
                 & & HasA & physicalPartOf & meronym (WN) & has part & \\
         \bf  part-whole       & & MadeOf & memberOf & holonym (WN) & member of &  \\
                 & & AtLocation* & substanceOf & &material used \\ \hline
         & & AtLocation* & location & & location \\
        \bf spatial        & & LocatedNear & spatial & & anatomical location \\ \hline
        \bf creation & & CreatedBy & & & creator \\ \hline
        & & ReceivesAction \\
        \bf utility & & UsedFor & hassynsetmember & using (FN) & used by \\
         & & CapableOf & activity & & use &  \\
         & & $\neg$NotCapableOf & participant & & uses \\ \hline
         & xIntent & CausesDesire \\
        & xWant & MotivatedByGoal \\
        \bf desire/goal & oWant & Desires \\
        & & $\neg$NotDesires & \\
        & & ObstructedBy \\ 
        \hline
        & & & shape \\
        & & & size \\
        \bf quality & & HasProperty & color & frame\_element (FN)  \\
        & & $\neg$NotHasProperty & taste\_property & polarity (SN) & color \\
        & xAttr & SymbolOf & temperature & & has quality \\ \hline
        \bf comparative & & & \it 6.3k relations \\ \hline
        & xNeed & HasFirstSubevent & & subframe (FN) \\
        & xEffect & HasLastSubevent & time & precedes (FN) \\
        \bf temporal & oEffect & HasSubevent & emotion & inchoative\_of (FN) \\
        & xReact & HasPrerequisite & prev & causative\_of (FN) & has cause &  \\
        & oReact & Causes & next & & has effect &  \\
        & & Entails & \\ \hline
        & & RelatedTo & &  & field of this occupation\\
        \bf relational & & HasContext & thing & see\_also (FN) & depicts\\
        \bf -other & & EtymologicallyRelatedTo & agent & requires (FN) & health specialty\\
		\bottomrule
	\end{tabular}}
}
\end{table*}

\section{Dimensions of commonsense knowledge}
\label{sec:dimensions}
In the previous section, we surveyed 20 representative commonsense sources from five categories: commonsense KGs, common KGs, lexical sources, visual sources, and corpora and language models. A key contribution of this paper is a manual categorization (by the authors) of the kind of knowledge expressed by the relations in these sources into 13 dimensions.
Table \ref{tab:dims} shows the correspondence of each relation in these analyzed sources to our dimensions. An example for each of the dimensions from different sources is shown in Table \ref{tab:examples}. We next describe each dimension in turn.

\textbf{Lexical.}  Many data sources leverage the vocabulary of a language or the lexicon in their relations. This includes relationships such as plural forms of nouns, or past tenses of verbs, for example. Lexical knowledge also covers substring information. ConceptNet, for example, includes a relationship called \texttt{DerivedFrom} that they describe as capturing when a word or phrase appears within another term and contributes to that term's meaning. Lexical knowledge is also the formalization of the relation between a concept and its expression in a language, e.g., denoted by through the \texttt{label} relation in Wikidata.

\textbf{Similarity.}  Most data sources include the notion of synonymy between expressions. They typically allow definitions of terms, and some cover a broader notion of general similarity. 
ConceptNet has all three subcategories - for instance, regarding similarity, it establishes that wholesome and organic food are similar notions, while eating is defined as process of taking in food. WebChild also captures similarity between WordNet concepts, while WordNet, Wikidata, and Roget focus on synonymy. For instance, Roget declares that food and edibles are synonyms, while Wikidata expresses that food is \textit{said to be the same as} nutriment.

\textbf{Distinctness.} Complementary to similarity, most data sources have notions of some kind of distinguishability. Most commonly, this is formalized as antonymy, where words have an opposition relationship between them, i.e., they have an inherently incompatible relationship. For example, both Roget and ConceptNet consider hot and cold to be antonyms, as these are two exclusive temperature states of objects. FrameNet defines an \texttt{Excludes} relation to indicate that two roles of a frame cannot be simultaneously filled in a given situation. For instance, in the \textit{Placing} frame, an event can either be brought by a cause event or by an intentional agent, but not both. Weaker forms of distinctness are defined by Wikidata and ConceptNet, for concepts that might be mistaken as synonyms. For example, Wikidata states that food safety is different from food security, while ConceptNet distinguishes food from drinks.

\begin{table*}[!t]
	\centering
	{
%	\small
	\label{tab:examples}
	\caption{Examples for \textit{food} for each of the 13 dimensions. When the subject is different from food, we state it explicitly, e.g., \texttt{xWant: watch movie together - get some food}.}
\begin{tabular}{ | l | l | l | }
\hline
	\bf Dimension & \bf Example & \bf Source \\ \hline
	lexical & derivationaly related form: nutrient & WordNet \\ 
	 & etymologically related: fodder & ConceptNet \\ 
	 & derived term: foodie & ConceptNet \\ \hline
	similarity & synonym: dish  & ROGET \\
	 & said to be the same as: nutriment & Wikidata \\
	 & similar to: wholesome - organic & ConceptNet \\ \hline
	distinctiveness & opposite of: non-food item & Wikidata \\
	& distinct from: drink & ConceptNet \\
	& different from: food safety - food security & Wikidata \\ \hline
	taxonomic & hyponym: comfort food & WordNet \\ 
	 & hyponym: beverage & WordNet \\ 
	 & hypernym: substance & WordNet \\
	 & subclass of: disposable product & Wikidata \\ \hline
	part-whole & things with food: minibar & ConceptNet \\ 
	 & is part of: life & COMET \\
	 & material used: food ingredient & Wikidata \\ \hline
	spatial & is located at: pantry & ConceptNet \\ 
	 & is located at: a store & ConceptNet \\
	 & location: toaster - kitchen & Wikidata \\
	 & located near: plate & Visual Genome\\
	 & located near: table & Visual Genome \\ \hline
	creator & is created by: cook & COMET \\ 
	 & is created by: plant & COMET \\ \hline
	utility & use: eating & Wikidata \\ 
	 & used by: organism & Wikidata \\ 
	 & used for: pleasure & ConceptNet \\ 
	 & used for: sustain life & COMET \\ 
	 & used for: nourishment & ConceptNet \\
	 & capable of: cost money & ConceptNet \\ 
	 & capable of: go rotten & ConceptNet \\
	 & is capable of: taste good & COMET \\ \hline
	 goal/desire & xWant: watch movie together - get some food & ATOMIC \\ 
	 & desires: regular access to food & ConceptNet \\
	 & not desires: food poisoning & ConceptNet \\
	 & causes desire to: eat & ConceptNet \\ 
	 & xIntent: eats food - quit feeling hungry & ATOMIC \\ 
     & motivated by: cook a meal & ConceptNet \\ 
	 & is motivated by: you be hungry & COMET \\ \hline
	 quality & xAttr: makes food - creative & ATOMIC \\ 
	 & has quality: shelf life & Wikidata \\ 
	 & has the property: tasty & COMET \\ \hline
	comparative & healthier: home cooking - fast food & WebChild \\ \hline
	temporal & has first subevent: cooking & ConceptNet \\
	 & starts with: open your mouth & COMET \\
	 & has effect: food allergy & Wikidata \\ 
	 & causes: you get full & COMET \\
	 & causes: indigestion & COMET \\ \hline
	relational-other & related to:  refrigerator & ConceptNet \\
	& related to: cereal & ConceptNet  \\
	& field of work: food bank - food assistance & Wikidata \\
	& main subject: cuisine - food product & Wikidata \\ \hline
\end{tabular}
}
\end{table*}

\textbf{Taxonomic.} Most data sources include a kind of arrangement classification where some objects are placed into more general and more specific groupings with inheritance relations. When those groupings are ordered categories based on generality, this captures the notion of hyponymy, indicating a subcategory relationship. Hyponymy blends the distinction between the relationships \texttt{subclass} (intended for two classes) 
and \texttt{InstanceOf}  (intended as a relation between an instance and a class). For instance, Wikidata states that a sandwich wrap is street food, and that food is a disposable product. WordNet has information that beverage and comfort food are hyponyms of food. While this dimension generally focuses on concepts (nouns), it also includes a specialization relation for verbs. Here, the \texttt{MannerOf} relation in ConceptNet states that wheezing is a manner of breathing. Two of the sources we review in this paper, Probase~\cite{wu2012probase} and IsaCore~\cite{cambria2012semantic}, focus only on taxonomic relations.

\textbf{Part-whole.} Many data sources include a notion of being a part of or a member of something. Part-whole knowledge can be transitive, such as that of geographic containment, exemplified by New York City being a part of New York State, which is also part of the United states. Other part-of notions, such as \textit{member-of} are not necessarily transitive. A third category of part-whole knowledge is expressed with the material or the building blocks of an object, such as food being made of food ingredients. A useful distinction between these three notions of part-whole: physical part of (sunroof - car), member of (musician - duet), and substance of (steel - boiler), is provided by WebChild. The importance of this commonsense dimension is shown by HasPartKB~\cite{bhakthavatsalam2020dogs}, which is an entire resource dedicated to part-whole relations.

\textbf{Spatial.} Spatial relations describe terms relating to or occupying space. This may entail indicating the usual location of a concept, as in the \texttt{location} property in wikidata or the \texttt{AtLocation} in ConceptNet. ConceptNet expresses locations for geographic entities, for example Boston is at location Massachusetts, as well as for things that can contain things: butter is at location refrigerator. Similar to the latter case, Wikidata includes an example that toasters are located in kitchens. A weaker spatial relation is one of spatial proximity in WebChild or ConceptNet, specifying that, e.g., bikes are located near roads. While Visual Genome does not explicitly have a spatial relation, concepts occurring in the same image region can be represented with the \texttt{LocatedNear} relation~\cite{ilievski2020consolidating}. Example such statements include food being located near a plate or a table.

\textbf{Creation.}
This dimension describes the process or the agent that brought something into existence. ConceptNet gives an example that a cake is created by the bake process, COMET has information that food is created from plants, while Wikidata states that rifle factories create shotguns. Table~\ref{tab:dims} reveals that no other source has creation information.

\textbf{Utility}. This dimension covers a notion of fitness or usefulness of objects for some purpose.
ConceptNet's relation \texttt{UsedFor} expresses knowledge that `the purpose of A is B', with an example of food being used for pleasure or nourishment. 
Wikidata has several similar relations: \textit{use}, \textit{used by}, and \textit{uses}, which can express that platter is used for food presentation, or food is used by organisms.
ConceptNet includes the notion of \texttt{CapableOf}, described as `A is capable of B if A can typically do B', like food being capable of spoiling or ``going rotten'', or knives being capable of cutting.  Another related notion is that of receiving an action: a button may receive the push action.  While a button does not have the sole purpose of being pushed, it is capable of receiving that action, and by inference, it may respond to the action.

\textbf{Desire or goal.}  This dimension covers knowledge about an agent's motivation, i.e., their desires or goals. An agent may want to have something or wish for something to happen.  The agent typically has certain goals, aims, and/or plans, that may motivate or explain those desires.
The relation \texttt{Desires} in ConceptNet may indicate, e.g., that a person desires regular access to food. Its negated version, \texttt{NotDesires} expresses that a person does not desire poisoned food. ATOMIC has two relations: \texttt{xWant} and \texttt{oWant}, to indicate the desires of an agent or other agents in a given situation. For instance, when people watch a movie together, they want to get some food.
Regarding goals, ConceptNet includes the \texttt{MotivatedByGoal} and \texttt{ObstructedBy} relations to indicate the motivation and the constraint for a certain action. For instance, ConceptNet indicates that one's sleep is obstructed by noise, while COMET's extension of ConceptNet posits that people cook a meal because they are hungry.

\textbf{Quality}. Commonsense sources typically describe attributes of an agent or qualities related to an object. 
For example, ConceptNet and COMET include the relation \texttt{HasProperty}, to express knowledge like ice having the property cold and that food has property tasty.
ATOMIC uses \texttt{xAttr} to indicate that, for example, the person that cooks food often has the attribute hungry or creative. WebChild and Wikidata both provide more specific qualities, such as taste, temperature, shape, or color. For instance, WebChild would specify the plant color as green. SenticNet describes the perceived ``polarity'' of an object, e.g., cold food is generally seen as negative.

\textbf{Comparative.} WebChild performs comparison of objects based on relative values for their attributes. Example comparative relations in WebChild are: healthier than (home cooking - fast food), faster than (car - bike), and larger than (lion - hyena). Notably, no other source describes comparative knowledge explicitly.\footnote{Here we exclude implicitly comparative knowledge, such as the inferred information that eating food makes one more satisfied from the triple: PersonX eats food - xReact - satisfied. We also include a separate category for temporal representations that would include temporal comparison as one type of temporal information.}

\textbf{Temporal.} Most sources have notions of time that may support ordering by time and/or may capture relations that one thing is a prerequisite for another or one thing may have a particular effect.
ConceptNet, for example, expresses that the first event of eating may be cooking, while the last one could be getting rid of the containers. COMET states that eating starts with opening one's mouth. More strongly, the temporal relations often indicate relative ordering of two events, through relations of causation and effects, such as food potentially causing allergy or indigestion. Such causal knowledge is found in ATOMIC, ConceptNet, COMET, WebChild, and Wikidata.

\textbf{Relational-other.} Conceptual and context-related relationships are often underspecified. %this sentence was hard for me to parse.  i am rewriting.  
%On one hand, increasingly some sources capture description of the circumstances that for the setting for a statement, event, or idea. 
Increasingly resources are considering the setting or circumstances that surround an event or statement.
ConceptNet has a single relation \texttt{HasContext} for this, while Wikidata has more concrete contextual relations, such as \texttt{field of this occupation}, \texttt{depicts}, and \texttt{health specialty}. This allows Wikidata to express that the main subject of a cuisine is a food product, and that the field of work of food banks is food assistance.
Some resources have a kind of ``catch-all'' relationship that can be used very broadly. For example, 
most of the knowledge in ConceptNet belongs to a generic relation called \texttt{RelatedTo} that may be used to capture a relatively vague semantic connection between two concepts, such as food being related to refrigerator or cereal.

Our organization of existing relations into 13 dimensions provides a unified framework to reorganize and consolidate these sources. We also mention two nuances of our process. First, we placed the negative statements (marked with $\neg$ in Table~\ref{tab:dims}) in the same dimension as the positive ones, as they cover the same knowledge type, despite having a different polarity and, arguably, purpose. Following a similar line of reasoning, we also placed inverse relations, such as used for and uses, in the same dimension. Second, we recognize that the underlying data may not always be clearly placed in only one of these dimensions. For instance, the relation \texttt{AtLocation}, which intuitively should belong to the spatial category, contains some statements that express part-whole knowledge. 
\section{Experiments}
\label{sec:experiments}

Seven of the sources covered in the previous section: ConceptNet, ATOMIC, Visual Genome, WordNet, Roget, Wikidata-CS, and FrameNet, have been integrated together in CSKG~\cite{ilievski2020consolidating}. We start with CSKG and apply our dimension classification (section~\ref{sec:dimensions}) to its sources, under an assumption that each of their edge relations can be mapped unambiguously to exactly one of the dimensions that fits it the best.\footnote{As discussed before, this assumption might not always hold in practice. Future work should attempt to refine this mapping, e.g., by crowdsourcing or by clustering algorithms.} As a result, each edge in CSKG has dimension information stored in its \texttt{relation;dimension} column.\footnote{We leave out the relations prefixed with \texttt{/r/dbpedia} from ConceptNet, as these are being deprecated according to the official documentation: \url{https://github.com/commonsense/conceptnet5/wiki/Relations}.} CSKG contains knowledge for 12 out of our 13 dimensions - the dimension \textit{comparative} is not represented, as its only source, Web Child, is currently not part of CSKG. The resulting file is publicly available at: \url{http://shorturl.at/msEY5}.

This enrichment of the CSKG graph allows us to study the commonsense knowledge dimensions from multiple novel perspectives.
We investigate the following questions:
\begin{enumerate}
    \item \textit{Experiment 1:} How well is each dimension covered in the current sources? Here we compute the number of edges for each dimension across sources.
    \item \textit{Experiment 2:}  Is knowledge redundant across sources? We use the dimensions to quantify overlap between sources with respect to individual edges.
    \item \textit{Experiment 3:}  How do the edge dimensions compare to their lexical encoding?
We compute clusters based on our dimensions and compare them to clusters computed with Transformer-based language models, like BERT~\cite{devlin2018bert} and RoBERTa~\cite{liu2019roberta}.
    \item \textit{Experiment 4:} What is the impact of each dimension on downstream reasoning tasks?
Each of the dimensions is used to select a subset of the available knowledge in CSKG. The selected knowledge is then used to pretrain a RoBERTa language model, which is applied to answer commonsense questions in a zero-shot manner.
\end{enumerate}

In this section, we formulate and run suitable studies for each of the four questions, and reflect on the results.

\begin{table*}[!t]
	\centering
	{
%	\small
	\caption{Coverage of sources in terms of the knowledge dimensions. The numbers presented are in thousands.}
	\label{tab:coverage}

	\begin{tabular} {c | r r r r r r r r}
		\toprule
		\bf Dimension  & \bf ATOMIC & \bf ConceptNet & \bf WebChild & \bf Roget & \bf Wikidata-CS & \bf WordNet & \bf FrameNet \\	
		\midrule

        \bf lexical & & 704 & & & 0.5 & 207 & 14 \\ 
       \bf similarity   & & 255 & 343 & 1,023	& 1 &	152 & 0.4 \\
            \bf distinctness         & & 22 & & 381 & 7	& 4 \\ 
            \bf taxonomic     & & 244 &	783 & & 73 & 89 & 23 \\
         \bf  part-whole       & & 19 & 5,752 & & 8 & 22 \\
        \bf spatial        & & 28 & 660 & & 0.5 \\
        \bf creation & & 0.3 & & & 0.2 \\ 
        \bf utility & & 69 & 2,843 & & 2 & & 1 \\  
        \bf desire/goal & 244 & 20 \\
        \bf quality & 143 & 9 & 6,510  & & 1 & & 11 \\
        \bf comparative & & & 813 \\ 
        \bf temporal & 346 & 71 & 2,135 & & 3 & & 0.6 \\
        \bf relational-other & & 1,969 & 291 & & 6 & & 0.7 \\
		\bottomrule
	\end{tabular}
}
\end{table*}

\subsection{Experiment 1: How well is each dimension covered in the current sources?}
\label{ssec:exp1}

We use the CSKG graph enriched with edge dimensions to compute source coverage with respect to each dimension. The coverage of each source, formalized as a count of the number of edges per dimension, is presented in Table \ref{tab:coverage}.\footnote{Python script: \url{https://github.com/usc-isi-i2/cskg/blob/master/consolidation/compute_dimensions.py}.}

We observe several trends in this Table. First, we recognize imbalance between the number of sources per dimension. Comparative knowledge and creation information are very rare and are described by only one or two sources, whereas taxonomic, temporal, and similarity knowledge are much more common and are captured by most sources. Second, some of the dimensions, like creation or part-whole, are represented with relatively few edges, whereas similarity and taxonomic knowledge generally have a much larger number of edges. The exception for the former is the large number of part-whole statements in WebChild, which is due to the fact that WebChild is automatically extracted, resulting in many duplicates and noisy information. Third, we see that some sources, like ConceptNet, FrameNet, and Wikidata-CS, aim for breadth and cover most dimensions. Others, like Roget and ATOMIC, have a narrow focus on specific dimensions: primarily desires/goals and temporal knowledge in ATOMIC, and only knowledge on similarity and distinctness in Roget. Yet, the narrow focus generally coincides with much depth, as both sources have many edges for the small set of dimensions that they cover. FrameNet, having a broad focus, has a small number of edges for each dimension due to its limited coverage of lexical units. Again here, WebChild is a notable outlier with a large number of automatically extracted statements for most dimensions. Finally, we observe different ratios between `strong' and `weak' semantic relations across sources. Most of ConceptNet's knowledge falls under the generic \textit{relational-other} category, whereas only a small portion of Wikidata-CS belongs to the same dimension. Most of Wikidata-CS is taxonomic knowledge.

\begin{table}[!t]
	\centering
	{
	\caption{Overlap between various source pairs, based on the original relations (\texttt{Relations}) or the abstracted dimensions (\texttt{Dimensions}). Absolute overlap numbers are accompanied in brackets by the Jaccard percentage of the overlap against the union of all triples in the two sources.}
	\label{tab:overlap}

	\begin{tabular} {c | r r}
		\toprule
		\bf Source pair  & \bf Relations & \bf Dimensions \\	
		\midrule
		\bf CN - RG & 57,635 (1.23\%) & 73,992 (1.60\%) \\
        \bf CN - WD & 2,386 (0.07\%) & 2,623 (0.08\%) \\
        \bf CN - WN & 86,006 (2.14\%) & 97,946 (2.60\%) \\
        \bf RG - WD & 299 (0.02\%) & 333 (0.02\%) \\
        \bf RG - WN & 75,025 (3.55\%) & 75,025 (3.93\%) \\
        \bf WD - WN & 1,697 (0.19\%) & 1,704 (0.25\%) \\
		\bottomrule
	\end{tabular}
}
\end{table}

\begin{table*}[!ht]
	\centering
	{
	\caption{Overlap distribution across dimensions. Absolute overlap numbers are accompanied in brackets by the Jaccard percentage of the overlap against the union of all triples in the two sources. '-' indicates that at least one of the sources does not use the dimension.}
	\label{tab:overlap_dim}

    \scalebox{0.97}{
	\begin{tabular} {c | r r r r r r r r r r}
		\toprule
		\bf Sources  & \bf part-whole & \bf taxonomic & \bf lexical & \bf distinctness & \bf similarity & \bf quality & \bf utility & \bf creation & \bf temporal & \bf rel-other \\	
		\midrule
		\bf CN-RG &- &- &- & 4,639 & 69,353 & - & - & - & - & - \\
		&- &- &- & (1.17) & (5.79)& - & - & - & - & - \\ \midrule
        \bf CN-WD & 68 & 1,888 & 20 & 266 & 102 & 0 & 14 & 0 & 1 & 264 \\
        & (0.25) & (0.62) & (0.00) & (1.00) & (0.04) & (0.00) & (0.02) & (0.00) & (0.00) & (0.01) \\ \midrule
        \bf CN-WN & 4,710 & 73,123 & - & 1,053 & 19,060 & - & - & - & - & -   \\
        & (4.10) & (15.19) & - & (4.65) & (5.05) &- & -&- & - & -  \\ \midrule
        \bf RG-WD &- & -&- & 206 & 127 & - & - & - & - & - \\
        &- & -&- & (0.05) & (0.01) & - & - & - & - & - \\ \midrule
        \bf RG-WN &- & -&- & 3,300 & 71,725  & - & - & - & - & - \\
        &- & -&- & (0.87) & (6.50) & - & - & - & - & -   \\ \midrule
        \bf WD-WN & 82 & 1,533 & - & 63 & 26 & - & - & - & - & -   \\
        & (0.07) & (0.39) & - & (0.62) & (0.02) & - & - & - & - & -   \\
		\bottomrule
	\end{tabular}
	}
}
\end{table*}

\subsection{Experiment 2: Is knowledge redundant across sources?}

Our analysis so far reveals that most dimensions are covered by more than one source. This leads us to the next question: how often is a statement found in multiple sources?

Computing edge overlap between sources is conditioned on identity mapping between their nodes and relations.
While CSKG provides such identity mappings between some of its nodes, this cannot be expected to be complete. We align the edges as follows.
The \textbf{nodes} across sources are naively compared through their labels.\footnote{If a node has more than one label, then we perform comparison based on the first one.} 
Regarding the \textbf{relations}, we benefit from CSKG's principle of normalizing the relations across sources to a shared set. With this procedure, a WordNet edge \texttt{(food.n.01, synonym, dish.n.01)} is modelled as \texttt{(food, /r/Synonym, dish)} in CSKG. 
As a dimension-based enhancement, we abstract each relation further by mapping it to our dimensions, e.g., transforming \texttt{(food, /r/Synonym, dish)} to \texttt{(food, similarity, dish)}. This dimension-based transformation allows for more flexible matching within a dimension, for instance, enabling similarity and synonymy statements to be compared for equivalence, since both \texttt{(food, /r/Synonym, dish)} and \texttt{(food, /r/SimilarTo, dish)} would be normalized to \texttt{(food, similarity, dish)}.

We apply the relation-based and dimension-based variants to compute overlap between four sources: ConceptNet, Roget, Wikidata, and WordNet, in terms of each dimension. Here we do not consider ATOMIC or FrameNet, as their edges can be expected to have extremely low lexical overlap with the other sources. The overlap is computed as a Jaccard score between the number of shared triples between two sources and the union of their triples. The obtained scores are given in Table \ref{tab:overlap}.\footnote{Notebook: \url{https://github.com/usc-isi-i2/cskg/blob/master/analysis/Overlap.ipynb}.} We observe that the overlap is generally low, yet, translating the original relations into dimensions constantly leads to an increase of the overlap for any of the source pairs. The highest relative overlap is observed between Roget and WordNet (3.93\%), with ConceptNet-WordNet coming second (2.60\%). The lowest overlap is obtained between the sources Roget and Wikidata (0.02\%).

Next, we inspect the overlap between these sources for each dimension. Will the edges that correspond to more commonly found dimensions (e.g., part-whole, cf. Experiment 1) occur more often in multiple sources? We provide insight into this question in Table \ref{tab:overlap_dim}. Primarily, this Table reveals that there is very little edge overlap across sources. As hypothesized, most of the shared edges belong to dimensions that are common in many commonsense sources, describing \texttt{taxonomic}, \texttt{similarity}, and \texttt{part-whole} knowledge. The highest Jaccard score is obtained on the taxonomic knowledge between ConceptNet and WordNet, followed by similarity knowledge in ConceptNet-Roget and Roget-WordNet.
Wikidata and ConceptNet share edges that belong to a number of other dimensions, including \texttt{distinctness}, \texttt{similarity}, and \texttt{rel-other}. 

The sparse overlap in Tables~\ref{tab:overlap}-\ref{tab:overlap_dim} is amplified by our lexical method of computing overlap, as the same or similar nodes may have slightly different labels. Both the low overlap and the relatively weak comparison method strongly motivate future work on node resolution of commonsense KGs. Node resolution of commonsense KGs is non-trivial, as nodes are intended to represent various aspects of meaning: words, phrases, concepts, frames, events, and sentences. We discuss further the challenges and opportunities relating to node resolution of commonsense KGs in Section~\ref{sec:discussion}.

\subsection{Experiment 3: How do the edge dimensions compare to their lexical encoding?}
\label{ssec:exp3}

Next, we investigate how the information captured by our dimensions relates to the encoding of edges by state-of-the-art Transformer-based language models, like BERT or RoBERTa.\footnote{Notebook: \url{https://github.com/usc-isi-i2/cskg/blob/master/embeddings/Summary\%20of\%20Dimension\%20on\%20CSKG.ipynb}}
For this purpose, we cluster the knowledge in CSKG according to our 13 dimensions, resulting in 13 disjoint clusters. We also compute clusters based on language models in an unsupervised manner as follows. Each of the edges is lexicalized into a natural language sentence by relation-specific templates. Each sentence is then encoded with a Transformer model, either BERT-large or RoBERTa-large, into a single 1,024-dimensional embedding. These embeddings are finally clustered with the k-Means~\cite{hartigan1979algorithm} algorithm into $k=13$ disjoint clusters.

The two approaches for computing clusters, based on our dimensions and based on Transformer embeddings, can now be compared in terms of their agreement. We use the adjusted rand index (ARI) metric to measure agreement.\footnote{\url{https://scikit-learn.org/stable/modules/generated/sklearn.metrics.adjusted_rand_score.html}} The ARI score is 0.226 for BERT and 0.235 for RoBERTa. These scores signal low agreement between the dimension-based and the unsupervised clustering, which is expected given that the dimension-based clustering entirely depends on the relation, while the unsupervised clustering considers the entire triple. We also observe that the ARI score of RoBERTa is slightly higher, which might indicate that the relation has higher impact on the embedding in RoBERTa than in BERT.

\begin{figure}[!h]
    \centering
    \includegraphics[width=0.5\textwidth]{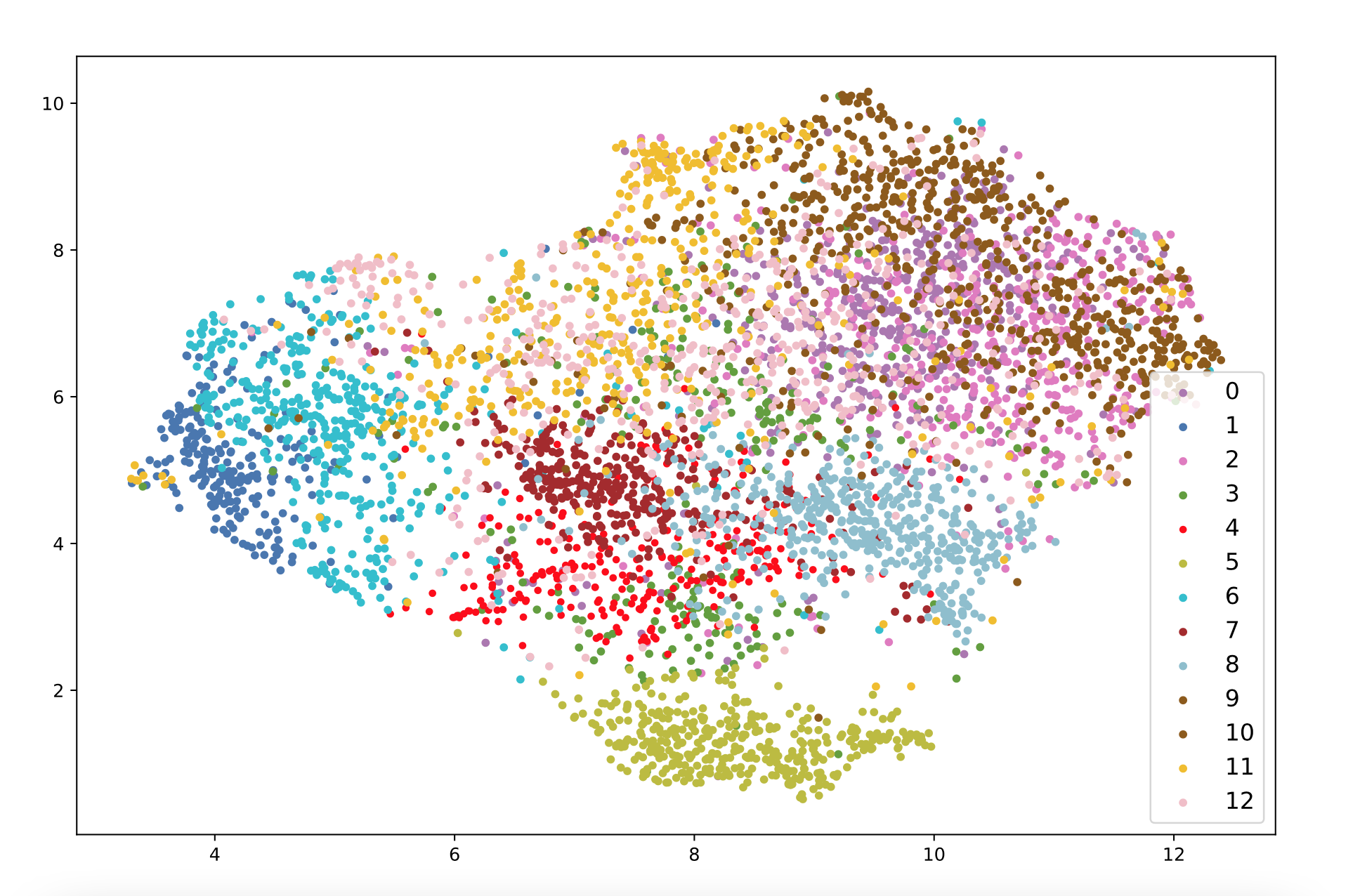}
    \caption{UMAP clusters of RoBERTa.}
    \label{fig:roberta_clusters}
\end{figure}

In order to understand the information encoded by RoBERTa, we use UMAP to visualize its k-means clusters for a random sample of 5,000 edges (Figure~\ref{fig:roberta_clusters}).
Curiously, certain clusters are clearly delineated, while others are not. For instance, cluster 5 has little overlap with the other clusters. Looking into the contents of this cluster, we observe that it is largely dominated by distinctness information: 92\% (360 out of 390) of its edges belong to this dimension, mostly expressed through the \texttt{/r/Antonym} relation. Clusters 4, 7, and 8 are largely dominated by similarity, while clusters 1 and 6 are largely split between temporal (46\%) and desire/goal (36\%) edges. At the same time, we observe a lot of overlaps between the clusters 0, 2, 9, 10, 11, and 12. These clusters are dominated by lexical and relational-other edges - e.g., around half of all edges in clusters 0 and 9 belong to the category relational-other. The node frequency distributions reveal that cluster 1 describes positive emotions, as its most frequent node is \text{/c/en/happy}; nodes in cluster 5 are often numbers, like \texttt{rg:en\_twenty-eighth}; and the nodes in cluster 9 describe concepts from natural sciences, the most connected node being \texttt{/c/en/zoology}.

\begin{figure}[!h]
    \centering
    \includegraphics[width=0.5\textwidth]{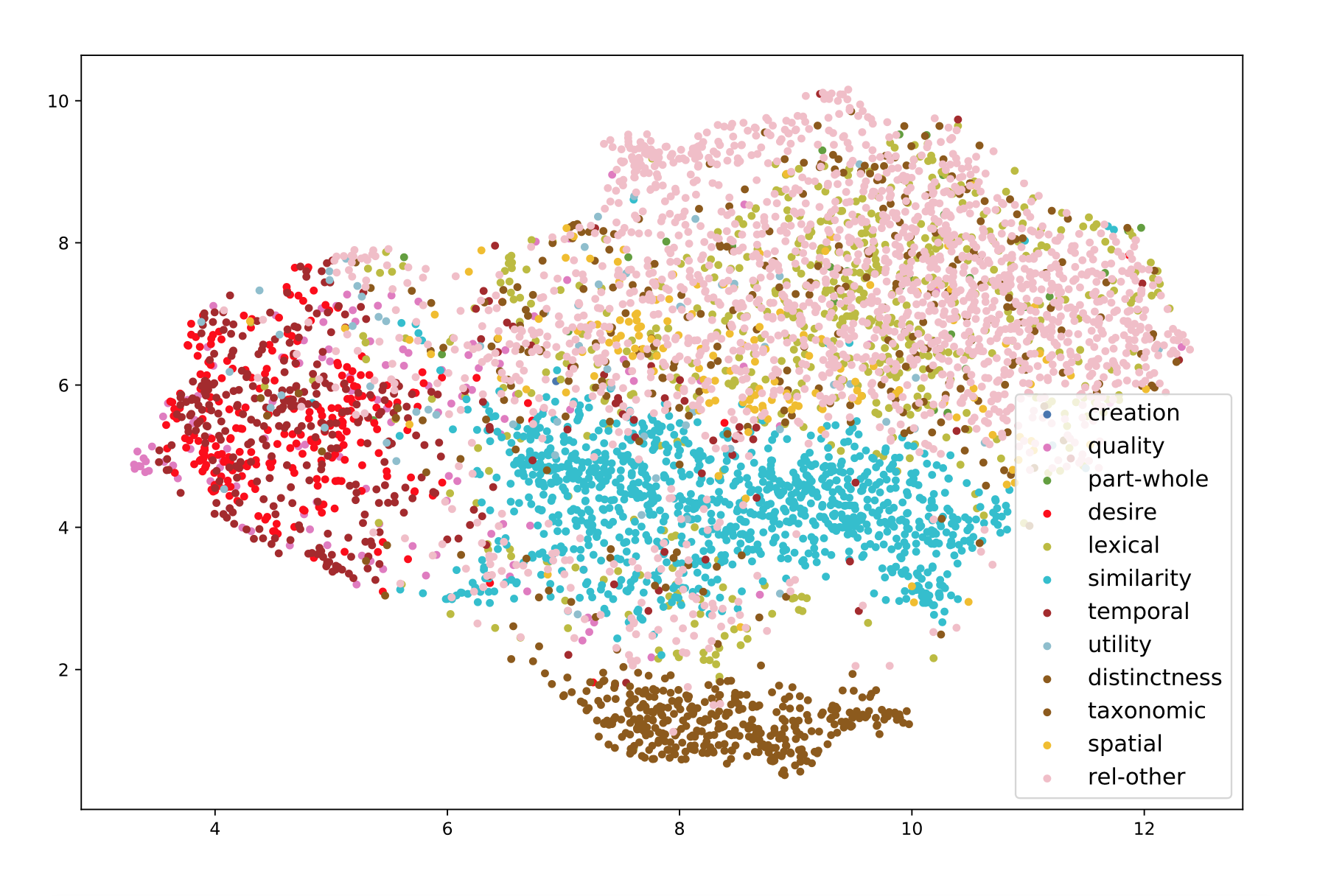}
    \caption{UMAP clusters according to our dimensions.}
    \label{fig:dim_clusters}
\end{figure}

\begin{table}[]
    \centering
    \caption{10 pairs with the highest Jaccard scores between dimension-based and RoBERTa-based clusters.}
    \label{tab:jaccard}

    \begin{tabular}{c c r} \toprule
    \bf RoBERTa cluster & \bf Dimension cluster & \bf Jaccard \\ \midrule
5	& distinctness &	0.916\\
8	& similarity	& 0.452\\
6	& temporal	& 0.322\\
7 &	similarity	& 0.295\\
6	& desire/goal	& 0.283\\
1	& desire/goal	& 0.258\\
4	& similarity &	0.210\\
0	& lexical	& 0.205\\
1	& temporal	& 0.202\\
12	& relational-other	& 0.182  \\ \bottomrule
\end{tabular}
\end{table}

\begin{table}[]
    \centering
    \caption{Top-3 highest-scored dimensions for each of the automatically-computed clusters. Bold-faced results indicate the top score for each dimension.}
    \label{tab:jaccard_top3}
    \begin{tabular}{c | r} \toprule
0 & \textbf{lexical (0.205)}, rel-other (0.121), taxonomic (0.060)\\
1 &  \textbf{desire (0.258)}, temporal (0.202), quality (0.087)\\
2 & lexical (0.133), rel-other (0.122), taxonomic (0.075)\\
3 & \textbf{spatial (0.119)}, lexical (0.061), quality (0.053)\\
4 & similarity (0.21), quality (0.015), lexical (0.008)\\
5 & \textbf{distinctness (0.916)}, lexical (0.018), taxonomic (0.003)\\
6 & \textbf{temporal (0.322)}, desire (0.283), quality, (0.054)\\
7 & similarity (0.295), quality (0.009), taxonomic (0.005)\\
8 & \textbf{similarity (0.452)}, lexical (0.004), taxonomic (0.003)\\
9 & rel-other (0.143), lexical (0.085), \textbf{taxonomic (0.081)}\\
10 & rel-other (0.169), taxonomic (0.015), quality (0.007)\\
11 & rel-other (0.11), \textbf{quality (0.068)}, taxonomic (0.066)\\
12 & \textbf{rel-other (0.183)}, taxonomic (0.059), spatial (0.053)\\
\bottomrule
\end{tabular}

\end{table}

In Figure~\ref{fig:dim_clusters}, we visualize the same set of edges, only this time we color each according to their dimension. In accordance with the relatively low rand index score, we observe that the clusters are mostly not well-distinguished from one another. We look for correspondences between the RoBERTa clusters in Figure~\ref{fig:roberta_clusters} and the dimension-based clusters in Figure~\ref{fig:dim_clusters}, by computing Jaccard score between the edges that constitute each pair of their clusters. The 10 cluster pairs with the highest Jaccard scores are shown in Table~\ref{tab:jaccard}. We observe the highest correspondence for distinctness with cluster 5 (Jaccard score of 0.92) and similarity with cluster 8 (score 0.45). The overlapping clusters for desire and temporal knowledge both map relatively strongly to clusters 1 and 6, which appear nearby in the RoBERTa clustering as well. We also observe relatively high score for the cluster pairs 4-similarity, 7-similarity, and 0-lexical, all of which confirm our prior analysis of Figure~\ref{fig:roberta_clusters}.

We show the top-3 highest-scored dimensions for each of the automatically-computed clusters in Table~\ref{tab:jaccard_top3}. Here, we observe that some dimensions, like \texttt{creation}, \texttt{utility}, and \texttt{part-whole} do not fall within the top-3 dimensions for any of the RoBERTa clusters. This is likely due to the lower number of edges with these dimensions, as well as their dispersion across many RoBERTa clusters. 

\begin{figure}[!h]
    \centering
    \includegraphics[width=0.5\textwidth]{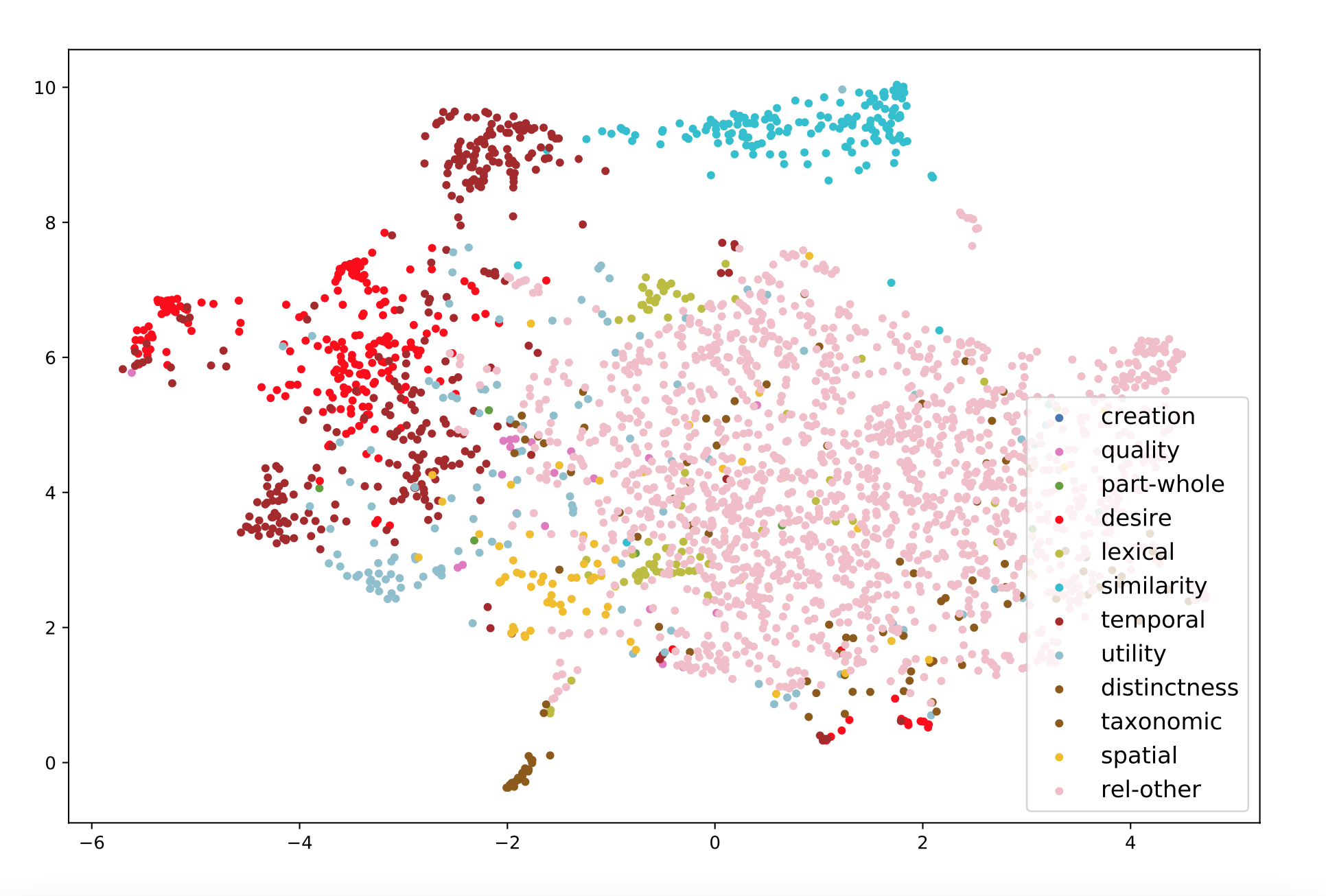}
    \caption{UMAP clusters for two selected nodes: \texttt{/c/en/food} and \texttt{/c/en/eat}.}
    \label{fig:two_node_clusters}
\end{figure}

To investigate further, we select the two CSKG nodes we considered earlier in Table~\ref{tab:examples}: \texttt{/c/en/food} and \texttt{/c/en/eat}, and visualize their edges according to their dimension. The total set of 2,661 edges largely belongs to the dimension \textit{relational-other} (1,553), followed by \textit{temporal} (319 edges), and \textit{desire/goal} (228 edges), whereas no edge belongs to the \textit{creation} dimension. Thus, most of the well-specified edges about nutritional concepts express either temporal information about the process, or knowledge about desires/goals relating to nutrition.
Within a single dimension, most \textit{relational-other} edges are expressed with \texttt{RelatedTo} (1,488 out of 1,553). \textit{Temporal} knowledge is split into multiple relations, primarily \texttt{HasLastSubevent} (113 edges), \texttt{HasPrerequisite} (69), and \texttt{HasSubevent} (60). \textit{Desire/goal} divided into \texttt{at:xWant} (78 edges), \texttt{MotivatedByGoal} (47), and \texttt{at:xIntent} (47 edges). 
Besides the two seed nodes (\texttt{/c/en/food} and \texttt{/c/en/eat}), the frequency distribution of nodes in a cluster reveals other prominent nodes. Naturally, the spatial dimension includes \texttt{/c/e/plate} (with an edge degree of 3), the temporal cluster includes \texttt{/c/en/diminish\_own\_hunger} (degree of 4), and the distinctness cluster has 7 edges for \texttt{/c/en/drink}.
%\dlm{ As i am reading these experiments i am wondering if there is an upshot kind of take home message that we can put at the beginning or end of each experiment or at the end of the experiments section overall}
%would be nice to have a summary of the answers to the driving experiment questions somewhere - in a last paragraph in this section might work - we KIND of have that in the in summary paragraph but it does not really go experiment by experiment or question by question

In summary, experiment 3 reveals a relatively low agreement between the dimensions and the language model-based clustering. This indicates that the nodes of an edge, which are not considered by the dimensions, are important for the language models. Meanwhile, we observe individual correspondences between the language models and our dimensions, most notably on the dimensions of similarity and distinctness. In addition, both the dimensions and the language models naturally cluster nodes from different sources together.

\subsection{Experiment 4: What is the impact of each dimension on downstream reasoning tasks?}
\label{ssec:reasoning}

Experiment 3 revealed overlaps between the information captured by our dimensions and that captured by language models. In our fourth experiment, we experiment with enhancing language models with knowledge belonging to individual dimensions, in order to examine the effect of different dimensions of knowledge on commonsense reasoning tasks. 

We adopt the method proposed by Ma et al.~\cite{ma2020knowledgedriven} to pretrain state-of-the-art language models and conduct zero-shot evaluation on two commonsense question answering tasks. According to this method, we first transform ConceptNet, WordNet, Wikidata, ATOMIC, and Visual Genome into synthetic QA sets. We use templates to map each triple into a QA pair and apply random sampling with heuristic-based filtering to collect two distractors for every QA pair. We group synthetic QA pairs based on their dimension, resulting in 12 dimension-based QA buckets in total. Within each dimension, the QA data is split into a training and development set. For ATOMIC, we adopt its original split to partition data. For the other sources, we select 95\% of the data as a training set and the remaining 5\% as a development set, following the original paper. We note that Ma et al.~\cite{ma2020knowledgedriven} only selected 14 relations from ConceptNet, WordNet, Wikidata, and Visual Genome, whereas we include all relations except \texttt{RelatedTo}. The statistics for the synthetic QA sets are shown in Table~\ref{tab:stats_dim}. We can see that the knowledge distribution across dimensions is fairly skewed, with creation having very few questions, while taxonomic and temporal knowledge being very well populated. Our experiments in this section will reveal whether the amount of available knowledge affects downstream task performance.  

\begin{table}[!t]
	\centering
	{
	\caption{Statistics of the number of QA pairs for each dimension.}
	\label{tab:stats_dim}
	\begin{tabular} {c | r r}
		\toprule
		\bf Dimension  & \bf Train & \bf Dev \\	
		\midrule
		\bf part-whole & 87,765 & 4,620 \\
        \bf taxonomic & 340,609 & 17,927 \\
        \bf lexical & 107,861 & 5,677 \\
        \bf distinctness & 20,286 & 1,068 \\
        \bf similarity & 166,575 & 8,768 \\
        \bf quality & 116,593 & 12,492 \\
        \bf utility & 63,862 & 3,362 \\
        \bf creation & 304 & 17 \\
        \bf temporal & 312,628 & 31,587 \\
        \bf relational-other & 242,759 & 12,777 \\
        \bf spatial & 21,726 & 1,144 \\
        \bf desire/goal & 194,906 & 20,912 \\
		\bottomrule
	\end{tabular}
}
\end{table}

We pretrain the RoBERTa-large \cite{liu2019roberta} model on each of the dimensions using the corresponding synthetic QA set. We use RoBERTa with a marginal ranking objective, as this is the best combination according to \cite{ma2020knowledgedriven}. 
We use the same set of hyper-parameters as in \cite{ma2020knowledgedriven}, except for the creation dimension. Specifically, we train our models for $1$ epoch using a learning rate of $1e-5$, a batch size of $32$, and $1.0$ as a margin. For the creation dimension, since the number of samples is much smaller, we train the model for $5$ epochs while keeping the other hyper-parameters fixed.  
We evaluate our models on two tasks: the CommonSenseQA (CSQA) task \cite{talmor-etal-2019-commonsenseqa}, in which the model is asked to choose the correct answer for a given question out of five options, and the SocialIQA (SIQA) task \cite{sap-etal-2019-social}, in which the model chooses the correct answer from three options based on a question and a brief context. 

The results from our experiments are shown in Table~\ref{tab:results}. 
Overall, we see that the zero-shot pretraining on any dimension allows the model to outperform the no-knowledge baseline. However, the variance of the improvement across dimensions is relatively large, revealing that certain dimensions are more relevant for the two downstream tasks than others. For example, although the training set size of \textit{lexical} dimension exceeds 107K, its performance gain on both tasks is limited. This is not surprising, since the language model can be expected to already have learned most of the lexical knowledge during its pretraining on unstructured text corpora. While the \textit{quality} dimension has a similar training set size as the \textit{lexical} dimension, the model benefits from it by a large margin on both tasks: 20.7 and 12.7 absolute points, respectively. This finding suggests that learning additional knowledge on object qualities is novel and useful, as it has not been fully acquired from unstructured data. 

We note that reasoning on downstream tasks benefits especially from the knowledge dimensions that align with the task question types. For example, each question in SIQA corresponds to an ATOMIC relation, and requires knowledge primarily about the order of events, personal attributes, and agent desires. Consequently, pretraining on \textit{quality}, \textit{temporal}, and \textit{desire/goal} knowledge provides the model with the largest gain on the SIQA task. The improvements of the \textit{temporal} dimension are even higher than training on the entire set of questions, suggesting that certain knowledge dimensions that are not related to social reasoning may even lead to a decline in model performance. For the CSQA task, since it is derived from the broad set of knowledge dimensions covered in ConceptNet, we expect that many (if not all) of these dimensions would help performance. Accordingly, we observe large gains with many of the knowledge dimensions (+15\%), whereas training on the utility dimension yields the best performance (+22.4\%), even slightly better than training on the entire set of questions. 

\begin{table}[!t]
	\centering
	{
	\caption{Zero-shot evaluation results on two commonsense reasoning tasks. We run every experiment 3 times with different seeds, and report mean accuracy with a 95\% confidence interval.}
	\label{tab:results}
	\begin{tabular} {l | c c}
		\toprule
		\bf Dimension  & \bf CSQA & \bf SIQA \\	
		\midrule
		\bf Baseline & 45.0 & 47.3 \\
		\bf +part-whole & $63.0 (\pm 1.4)$ & $52.6 (\pm 1.9)$ \\
        \bf +taxonomic & $62.6 (\pm 1.4)$ & $52.2 (\pm 1.6)$ \\
        \bf +lexical & $49.9 (\pm 2.9)$ & $49.0 (\pm 0.4)$ \\
        \bf +distinctness & $57.2 (\pm 0.5)$ & $50.2 (\pm 1.5)$ \\
        \bf +similarity & $61.4 (\pm 0.8)$ & $53.5 (\pm 0.6)$ \\
        \bf +quality & $65.7 (\pm 0.5)$ & $60.0 (\pm 0.7)$ \\
        \bf +utility & $\bf 67.4 (\pm 1.0)$ & $54.8 (\pm 0.7)$ \\
        \bf +creation & $49.9 (\pm 1.1)$ & $47.8 (\pm 0.2)$ \\
        \bf +temporal & $67.3 (\pm 0.3)$ & $\bf 62.6 (\pm 0.9)$ \\
        \bf +relational-other & $58.2 (\pm 1.7)$ & $51.3 (\pm 1.7)$ \\
        \bf +spatial & $63.3 (\pm 0.2)$ & $53.1 (\pm 0.3)$ \\
        \bf +desire/goal & $65.0 (\pm 1.8)$ & $60.0 (\pm 0.6)$  \\
        \bf +all  & $66.2 (\pm 1.4)$ & $61.0 (\pm 0.7)$  \\
		\bottomrule
	\end{tabular}
}
\end{table}

\begin{figure*}
    \centering
    \includegraphics[width=1\textwidth]{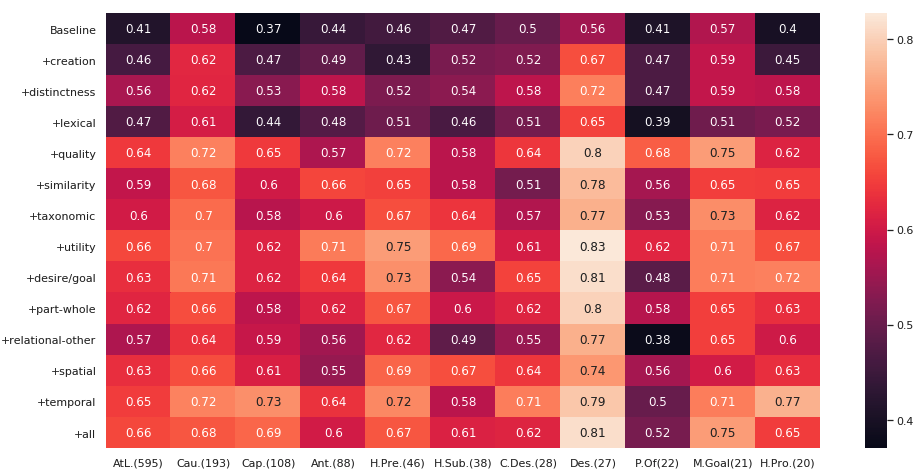}
    \caption{Accuracy for each question type in CSQA, where \textbf{AtL.} means AtLocation, \textbf{Cau.} means Causes, \textbf{Cap.} means CapabelOf, \textbf{Ant.} means Antonym, \textbf{H.Pre.} means HasPrerequisite, \textbf{H.Sub.} means HasSubevent, \textbf{C.Des.} means CauseDesires, \textbf{Des.} means Desires, \textbf{P.Of} means ParfOf, \textbf{M.Goal} means MotivatedByGoal, \textbf{H.Pro} means HasProperty. The numbers in parentheses indicate how many questions fall into the category.}
    \label{fig:csqa}
\end{figure*}

\begin{figure*}
    \centering
    \includegraphics[width=1\textwidth]{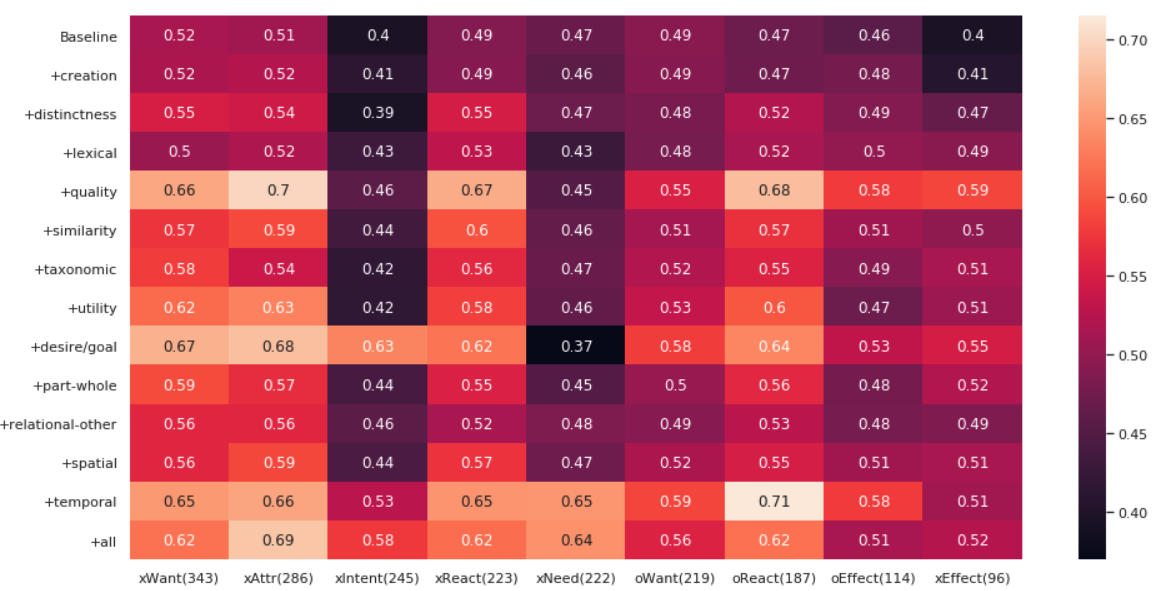}
    \caption{Accuracy for each question type in SIQA. The numbers in parentheses indicate how many questions fall into that category.}
    \label{fig:siqa}
\end{figure*}

To improve our understanding of the impact of every dimension of knowledge on different questions, we further break down the performance of the models per question type. For CSQA, we classify the questions based on the ConceptNet relations between the correct answer and the question concept, as the model needs to reason over such relations in order to be able to answer the question. For SIQA, since the questions are initially generated using a set of templates based on ATOMIC relations, we try to reverse-engineer the process by manually defining a mapping from the question format to the ATOMIC relations. Using this method, we are able to successfully classify more than 99\% of questions from SIQA's development set. For every question type, we compute the average accuracy over 3 seeds for the models trained on each dimension. The results on CSQA and SIQA are shown in the Figures \ref{fig:csqa} and \ref{fig:siqa}, respectively.\footnote{For CSQA, we omit question types with less than 20 questions.}

For some question types in CSQA, (one of) the largest improvements is achieved by training on the corresponding knowledge dimension. For example, \textit{temporal} knowledge for questions of type \texttt{Causes}, and \textit{desire/goal} for \texttt{Desires} questions. However, in other cases, the accuracy improvement by the corresponding knowledge dimension is much lower than other dimensions, for example, \textit{distinctness} knowledge is less useful for \texttt{Antonym} questions compared to \textit{utility} knowledge. This could be interpreted as a signal that the knowledge might not be clearly separated between the dimensions. For SIQA, the results show that the corresponding knowledge dimension is clearly impactful for most question types: \textit{desire/goal} on \texttt{xWant}, \texttt{xIntent}, \textit{quality} on \texttt{xAttr} and \textit{temporal} on \texttt{xNeed}, \texttt{oReact}, \texttt{oEffect}. This is especially visible for \texttt{xIntent} and \texttt{xNeed}, where very little gain is observed for most knowledge dimensions except for the corresponding dimension, suggesting that aligning the questions to their knowledge dimensions is important for building effective models. We note that a similar finding on the alignment between knowledge and the task has been reported in the original paper~\cite{ma2020knowledgedriven}; yet, the dimensions allow us to validate this claim more precisely. 

Finally, to verify our hypothesis that certain dimensions of knowledge are already learned by the state-of-the-art language models to a large extent, while others are not, we directly evaluate the baseline LM on the synthetic QA sets for each dimension. The results are shown in Table~\ref{tab:difficulty}. As expected, even without any training, the model already achieves a very high accuracy (over 90\%) on the \textit{lexical} dimension, thus it could not receive much training signal from this dimension. Meanwhile, the LM accuracy on \textit{quality}, \textit{temporal}, and \textit{desire/goal} dimensions is significantly lower. This is mostly because questions from ATOMIC take a large portion in these dimension. As reported in \cite{ma2020knowledgedriven}, the questions from ATOMIC are more challenging than those created from other knowledge resources. We note that the accuracy for \textit{relational-other} is the lowest among all dimensions. We hypothesize that this is because this dimension is noisier than others and the knowledge in this dimension is less likely to be found in the unstructured text. We leave further investigation of this result for future research.

In summary, we observe that certain dimensions of knowledge are very beneficial and novel for language models, allowing them to improve their performance on downstream reasoning tasks. Other dimensions, like lexical knowledge, are almost entirely redundant, as the language models have already acquired this knowledge during their initial training. The exact contribution for each dimension depends on the knowledge required by the task at hand. We discuss the implications of the obtained results further in Section~\ref{sec:discussion}.

\begin{table}[!t]
	\centering
	{
	\caption{Zero-shot evaluation results of RoBERTa on the synthetic QA sets.}
	\label{tab:difficulty}
	\begin{tabular} {l | c }
		\toprule
		\bf Dimension  & \bf Dev accuracy \\	
		\midrule
		\bf part-whole & 67.5  \\
        \bf taxonomic & 57.0 \\
        \bf lexical & 90.1 \\
        \bf distinctness & 77.3 \\
        \bf similarity & 65.6 \\
        \bf quality & 45.5 \\
        \bf utility & 67.9 \\
        \bf creation & 82.4 \\
        \bf temporal & 47.2 \\
        \bf relational-other & 37.6 \\
        \bf spatial & 56.9 \\
        \bf desire/goal & 48.0 \\
		\bottomrule
	\end{tabular}
}
\end{table}

\section{Related Work}
\label{sec:related_work}

\subsection{Epistemic Foundations}

World knowledge is what people learn and generalize from physical and social experiences, distilling mental representations out of the most significant aspects of everyday life.\footnote{According to the classic argument of the mind-body problem, it is inherently impossible to characterize how generalization occurs, due to an {\it explanatory gap} \cite{levine1983materialism}.} 
In these terms, commonsense knowledge can be conceived as the partition of world knowledge that is {\it commonly shared} by most people. This definition, however, has intrinsic limitations: in fact, the scale and the diversity of physical and social experiences, as well as the ecological, context-dependent nature of what constitutes `common' and `uncommon' \cite{gibson2000ecological}, make it hard to formulate any abstract criterion of what should fall under commonsense knowledge. 
From Aristotle’s theory of categories \cite{aristotle2012metaphysics} to Brentano's empirical psychology \cite{brentano2014psychology}, deriving knowledge dimensions from empirical observations, as opposed to from {\it abstract criteria} \cite{hicks1904idealism}, is a fundamental epistemic approach that contributed to the birth of Cognitive Science as a discipline \cite{miller2003cognitive}, besides serving as reference framework for our current investigation.
In this article, in fact, we neither propose nor adopt any {\it a priori} principle to define what should be included in a commonsense knowledge graph; rather, we analyze multiple knowledge graphs and, supported by empirical methods and experimental validations, elicit their most salient conceptual structures. \\
%dlm i really like this clear statement - we might make sure this comes out earlier in our intro as well
In the history of general knowledge base systems, the difficulty of characterizing common sense has been a driver, rather than an obstacle. For instance, Cyc~\cite{lenat1995cyc}, the most monumental effort to construct an axiomatic theory of commonsense knowledge, has been actively growing for almost forty years. At present, the Cyc knowledge base comprises around 1.5 million general concepts, and 25 million rules and assertions about these concepts. Different domain-specific extensions of Cyc exist, funded by industry and government programs: considering the key role that commonsense knowledge can play in enhancing AI systems for private and public enterprises, Cyc's strategy of steering the general knowledge base development in the direction of domain use cases can represent a sustainable business model for other stakeholders in the field. Since our focus is on computational resources of commonsense knowledge, it is beyond the scope of this paper to discuss seminal work on axiomatization of common sense, such as Pat Hayes' \textit{naive physics}~\cite{hayes}, Gordon and Hobbs' \textit{commonsense psychology}~\cite{gordon2017formal}, and Ken Forbus' work on \textit{qualitative commonsense reasoning}~\cite{forbus1997qualitative}. Similarly, although we recognize the important role that semantic discourse theory plays in describing  commonsense utterances based on expressive logical criteria \cite{asher1990commonsense,lascarides1993temporal}, it is beyond the scope of the current contribution to consider whether the knowledge resources we illustrated can be used as a basis to study the formal structures of dialog and, in general, socio-linguistic interaction, grounded on common sense.
\\
Existing commonsense knowledge graphs make implicit categorizations of knowledge, by defining a tractable set of relations which can be traced to some types proposed in cognitive research. For instance, WebChild’s~\cite{tandon2017webchild} part-of relations resemble the partonomic-metonimic relations in cognitive science literature (e.g., see \cite{casati1999parts, winston1987taxonomy}), while ConceptNet~\cite{speer2017conceptnet} defines 34 relations, where the relation \texttt{IsA} can be often approximated with taxonomic knowledge. In its first version~\cite{liu2004conceptnet}, ConceptNet defined 20 relations grouped into 8 categories: K-lines, Things, Agents, Events, Spatial, Causal, Functional, and Affective. Zhang et al.~\cite{zhang2020winowhy} extrapolate the types in the Conceptual Semantic Theory with those in ConceptNet 1.0 and propose the following six categories: property, object, eventuality, spatial, quantity, and others. According to \cite{maclachlan1994framing}, four types of contextual knowledge are essential for humans to interpret or frame in text: intratextual, intertextual, extratextual, and circumtextual knowledge. These generic knowledge types, which are orthogonal to the declarative/procedural distinction, are relevant for most natural language understanding tasks. In \cite{ilievski2017hunger}, we analyzed these types for the task of entity linking; when it comes to commonsense question answering, they might provide a guide for extending the coverage of the knowledge, when combined with specific theories/axioms, such as those defined by the resources in Section \ref{ssec:ckgont} or in Cyc.  \\
Beyond the structural differences, commonsense knowledge graphs share the same foundational elements.  Commonsense knowledge is generally split into declarative and procedural, where the former is contained in unconditional assertions, and the latter requires conditional assertions\footnote{The distinction between these types of assertions was formalized in a seminal work by Gentzen \cite{gentzen1934investigations}.}: we can state, for instance, that windows are {\it typically} made of glass (declarative), and assert that if a large rock is thrown against a window, the glass typically breaks (procedural). As the use of the adverb {\it typically} suggests, commonsense knowledge rules out exceptions from the context of interpretation: for instance, bulletproof glass doesn't break when hit by a rock. \\
There is a consensus that commonsense knowledge describes the world in terms of objects with their properties and relations. Yet, there is no consensus in terms of how this knowledge should be formalized and used for reasoning. Moore \cite{moore1982role} argues that reasoning over existentially quantified prepositions, negations, or by cases, all require an approach based on formal logic. State-of-the-art techniques for commonsense reasoning~\cite{lin2019kagnet,ma2019towards,ma2020knowledgedriven} are largely based on statistical (language) models, complemented with knowledge represented in a textual form~\cite{sap2019atomic,speer2017conceptnet}. \\
Statistical models are largely based on co-occurrence patterns and primarily capture contextual similarity of concepts, e.g., based on lexical co-occurrence and structured patterns~\cite{rohde2006improved}. Recent work has investigated the extent to which purely statistical models, such as the Transformer models, can perform reliable and generalizable inference. The work by \citet{ettinger2020bert} shows that BERT struggles with complex inference, role-based event prediction, and the contextual impacts of negation. \citet{bhagavatula2019abductive} reveal that pretrained language models perform relatively well on commonsense inference, except on tasks that require numeric and spatial understanding. The logical commonsense probes in RICA~\cite{zhou2020rica} show that LMs perform similar to random guessing in the zero-shot setting, they are heavily impacted by statistical biases, and are not robust to linguistic perturbations. This is in line with the findings of \citet{elazar2021back}, which indicate that the good performance of LMs mostly comes from fine-tuning on the task data. Recognizing that commonsense reasoning largely relies on pragmatic understanding~\cite{grice1975logic}, Jeretič et al.~\cite{Jeretic2020AreNL} demonstrate that BERT exhibits partial ability for pragmatic reasoning on certain scalar implicatures and presuppositions. \\
It has been shown that language models have acquired certain commonsense knowledge during their pre-training on large textual data~\cite{petroni2019language, davison-etal-2019-commonsense}.
This inspired follow-up work that queried/prompted language models directly, treating them as knowledge bases~\cite{shwartz-etal-2020-unsupervised,autoprompt:emnlp20}. Alternatively, language models can be pretrained with structured knowledge and applied to downstream reasoning tasks in a zero-shot manner~\cite{ma2020knowledgedriven,banerjee2020self}.
Considering the large capacity of the pre-trained language models and the goal of generalizing to unseen tasks~\cite{ma2020knowledgedriven}, lightweight alternatives to fine-tuning have been proposed recently. One approach is to preserve the knowledge acquired during pretraining, but complement it by updating a small amount of additional parameters~\cite{lin-etal-2020-exploring, li2021prefixtuning}. Alternatively, one could keep the entire model intact and tune the model input for optimal knowledge elicitation~\cite{jiang-etal-2020-know, autoprompt:emnlp20}. Ultimately, it is unclear whether aspects like negation and case-based reasoning can be fully solved by language models, or require inclusion of structured knowledge and formal logic, as suggested by Moore~\cite{moore1982role}.  \\

\subsection{Consolidation efforts}

In this paper, we analyzed individual knowledge graphs through suitable semantic dimensions, with the goal of providing insights on how alignment and consolidation
of commonsense knowledge resources can be guided and, eventually, achieved. A natural extension of our work would be to evaluate ongoing efforts that adopt alternative methods of consolidation: accordingly, Framester \cite{gangemi2016framester},  BabelNet \cite{navigli2012babelnet}, CSKG \cite{ilievski2020consolidating}, and Predicate Matrix \cite{de2016multilingual} constitute some of the most mature projects in this space. In Framester, several resources like  WordNet, VerbNet, Frame\-Net, and BabelNet, are aligned using an OWL schema based on \textit{Description and Situations} and \textit{Semiotics} ontology design patterns.\footnote{These patterns can be accessed at \url{http://ontologydesignpatterns.org/wiki/Main_Page}} CSKG, which we leverage in this paper, is also based on a schema, but it doesn't rely on traditional RDF/OWL semantics: in fact, CSKG is a hyper-relational graph represented in a tabular format, designed to preserve individual knowledge structures of resources like ConceptNet, WebChild, Visual Genome, etc., exploit direct mappings when available, derive indirect mappings when possible (e.g., while ConceptNet and Visual Genome do not have direct connections, they both have mappings to WordNet), and infer links through statistical algorithms. BabelNet is a multilingual lexicalized semantic network based on automatically linking Wikipedia with WordNet, and expanded by using additional information from resources like FrameNet and VerbNet. Finally, the Predicate Matrix exploits Word Sense Disambiguation algorithms to generate semi-automatic mappings between FrameNet, VerbNet, PropBank, WordNet, and the Events and Situations Ontology~\cite{segers2015eso}.\footnote{\url{https://github.com/newsreader/eso-and-ceo}}\\
\citet{ji2021survey} survey knowledge graphs in terms of their representation principles, acquisition methods, and applications. This work considers commonsense reasoning as a key emerging application area for knowledge graphs. A review of the state-of-the-art commonsense reasoning is provided by~\citet{Storks2019CommonsenseRF}, with a strong focus on benchmarks and reasoning techniques. Recently, several tutorials on commonsense knowledge and reasoning have been held, with focus on consolidation~\cite{iswctutorial}, neural model reasoning~\cite{Sap2020CommonsenseRF}, neuro-symbolic integration~\cite{aaaitutorial}, acquisition~\cite{tandon2018commonsense,razniewski2021information}, and broader applications~\cite{tandon2018commonsense}.
\section{Discussion and Roadmap}
\label{sec:discussion}

\subsection{Summary of Findings}
Commonsense knowledge sources use different levels of semantics, come in a variety of forms, and strive to capture diverse notions of common sense. After surveying 20 commonsense knowledge sources, we proposed that their relations can be grouped into 13 dimensions, namely: lexical, similarity, distinctness, part-whole, spatial, creation, utility, desire/goal, quality, comparative, temporal, and relational-other. Most relations can be unambiguously mapped to one of the dimensions. 
We apply our dimensions to reorganize the knowledge in the CSKG graph~\cite{ilievski2020consolidating}. Following our devised mapping of relations to dimensions, we add an additional column (\texttt{relation;dimension}) in CSKG, indicating the dimension of each edge. This allows us to make use of the consolidation of seven existing sources done by CSKG, and complement it with the dimensions in order to perform more abstract analysis of its knowledge types.
We designed and ran four experiments to analyze commonsense knowledge in CSKG through the lenses of these 13 dimensions.

In experiment 1, we investigated the coverage of the 13 dimensions in current sources. Some dimensions, like part-whole and similarity, are a subject of interest in most sources. Others, like comparative knowledge and knowledge on desires/goals, are rarely captured. Yet, the depth of knowledge on the less commonly represented relations is still high, as illustrated by the 244 thousand desire/goal edges in ATOMIC. Here we also observed that the breadth of focus varies notably across sources, as some (e.g., ConceptNet and Wikidata-CS) cover a wide range of relations, while others (e.g., ATOMIC or WordNet) have a narrower focus.

Experiment 2 posed the question of whether individual knowledge statements are redundant across sources. Our experiments with four sources indicated, with few exceptions, that only a tiny portion of all edges were shared between a pair of sources. This experiment points to a two-fold motivation for node resolution over commonsense sources. On one hand, node resolution is needed to increase the quality of computing overlap beyond the current lexical comparison of nodes. On the other hand, as the sources have generally complementary goals, it is likely that even with a more semantic computation the overlap will remain low. Node resolution is, thus, essential to consolidate different views of a node (concept) into a single representation.

In experiment 3, we cluster all edges in CSKG according to their dimension, and compare these clusters with $k=13$ clusters based on a language model encoding of each edge. We noted that the overall agreement between the dimensions and the language model-based clustering is relatively low, indicating that language models pay much attention to the edge nodes. However, individual correspondences were noted. Similarity and distinctness quite clearly dominated some of the RoBERTa-based clusters, while other clusters were consistently split between the dimensions of desire/goal and temporal knowledge. Interestingly, the clusters inferred from the RoBERTa embeddings often clustered nodes from different sources into a single cluster.

Finally, in experiment 4 we investigated the impact of the dimensions on a downstream reasoning task of commonsense question answering. We adopted a recent idea~\cite{ma2020knowledgedriven} for pretraining language models with knowledge graphs. The best-scoring model in this paper, RoBERTa-large with marginal ranking loss, was provided with knowledge from one of our dimensions at a time, and evaluated in a zero-shot manner on two benchmarks testing broad (CommonsenseQA) and social (SocialIQA) commonsense reasoning. The experiments showed that social commonsense reasoning clearly benefits from temporal, quality, and desire/goal knowledge, whereas the CommonsenseQA benchmark benefits from broad knowledge from all dimensions. Certain dimensions, such as lexical knowledge, were relatively uninformative, as it can be expected that such knowledge has been already acquired by the language models at their initial training stage. While the extent of knowledge plays a role, adding more knowledge is not always beneficial, as the task performance depends on the alignment between the dimensions and the task. This motivates further work on automatic alignment between the task questions and the our dimensions.  
%dlm adding another future work motivation
This also motivates future work that attempts to evaluate the value of additional content additions related to particular dimensions aimed at improving certain kinds of tasks.
%dlm i find this topic quite interesting.  after submission of this paper, we might consider what it takes to do this - and also what it takes to determine which dimensions have more value in certain contexts.  

\subsection{Outlook}

The goal of consolidating and applying commonsense knowledge is an ambitious one, as witnessed by decades of research on this topic. Our 13 dimensions are an effort to reorganize existing commonsense knowledge through unification of its knowledge types. We see this as a necessary, but not sufficient, step towards a modern and comprehensive commonsense resource. The dimensions could facilitate, or complement, several other aspects of this pursuit:

\noindent \textbf{1. Node resolution} While potentially controversial, the consolidation of commonsense knowledge relations into dimensions/knowledge types is achievable with careful manual effort. This is largely due to the relatively small number of relations in most current sources. Resolving nodes across sources is another key aspect of this consolidation, strongly motivated by experiment 2 of this paper. Sources capture complementary knowledge, whose combination is prevented by the lack of mappings between their nodes. As nodes are currently intended to represent various aspects of meaning: words, phrases, concepts, frames, events, sentences --- their consolidation is not obvious. Moreover, the number of unique nodes in most sources is on the order of many thousands or even millions~\cite{ilievski2020consolidating}, preventing it from being solvable by mere manual effort. Node resolution could be framed as a `static' disambiguation/clustering task, where each statement for a node is to be classified into one of its meanings, similar to~\cite{chen2011combining}. Here, the set of meanings can be either present (e.g., WordNet synsets) or dynamically inferred from the data. An alternative, `dynamic' approach is defer the node resolution to task time and perform it implicitly as a task of retrieving evidence from a knowledge source~\cite{lewis2020retrieval}. Another option is a combination of the static and the dynamic approaches.

\noindent \textbf{2. Coverage and boundaries} 
At present, it is difficult to estimate the completeness of commonsense knowledge sources. With the relations organized into dimensions of knowledge, we gain insight into the volume of knowledge that falls within each of the dimensions. An ideal node resolution would take us one step further, allowing us to detect gaps, i.e., understand which relevant facts are not represented by any of the sources. If nodes are resolved to an ontology like WordNet, one could leverage its taxonomy to infer new information. For instance, ConceptNet is at present unable to infer that if barbecues are held in outdoor places, they could be, by extension, be held in a park or someone's patio. In addition, a more semantic resource would allow us to define constraints over the knowledge and detect anomalies and contradictory knowledge, which can be argued to define the boundaries of the knowledge that can be obtained. It is reasonable that such boundaries exist, as commonsense knowledge is characterized by commonness of its concepts and restricted set of relations~\cite{ilievski2020commonsense}.
Further, organizing by dimensions also allows us to describe strengths (and/or weaknesses) of resources.  For example, a resource that has many partonomic relationships might be the first resource to consider using if a task requires part-whole reasoning.

\noindent \textbf{3. Generalizable downstream reasoning} As current large-scale commonsense sources are primarily text-based, they are lexicalized prior to their combination with language models, losing much of their structure. As this lack of structure prevents us from understanding their coverage and gaps, we are unable to measure their potential for downstream reasoning as a function of the available knowledge. It remains unknown to which extent a more complete source, organized around dimensions of commonsense knowledge, would be able to contribute to improving performance. Experiment 4 showed that there is correspondence between knowledge dimensions and question answering tasks, motivating automatic alignment between the two. Moreover, a comprehensive semantic source may inspire new neuro-symbolic reasoning methods, with potentially enhanced generalizability and explainability, opening the door for reliable commonsense services to be made available in the future.

\noindent \textbf{4. Evaluation and knowledge gaps} Experiment 4 showed that the potential of different dimensions for reasoning varies greatly and is largely dependent on the evaluation data. This finding is in line with~\cite{ma2020knowledgedriven}. The fact that certain dimensions consistently contribute little can be an indicator of gaps in current evaluation. Namely, dimensions like \textit{distinctness} and \textit{spatial} which currently contribute little or not at all are likely to be underrepresented in current evaluations. These gaps should ideally be addressed in the future by new benchmarks that will represent these missing dimensions. We note that our set of dimensions is based on the relations found in current popular commonsense sources. Hence, in this paper, we make an assumption that the knowledge types in these sources suffice (or at least previously sufficed) and can express the desired knowledge. The diversity of knowledge expressed by the \textit{relational-other} dimension, as pointed out also in~\cite{ilievski2020commonsense}, might be an indicator for additional, latent dimensions hidden behind the vagueness of this dimension.

\section{Conclusions}
\label{sec:conclusions}

At present, commonsense knowledge is dispersed across a variety of sources with different foci, strengths, and weaknesses. The complementary knowledge covered by these sources motivates efforts to consolidate them into a common representation. In this paper, we pursued the goal of organizing commonsense relations into a shared set of knowledge dimensions in a bottom-up fashion. Starting from a survey and analysis of the relations found in existing sources, we grouped them into 13 dimensions: lexical, similarity, distinctness, part-whole, spatial, creation, utility, desire/goal, quality, comparative, temporal, and relational-other. As each relation in these sources can be mapped to a dimension, we applied our method to abstract the relations in an existing consolidated resource: the Commonsense Knowledge Graph. This allowed us to empirically study the impact of these dimensions. First, we observed that some dimensions are included more often than others, potentially pointing to gaps in the knowledge covered in existing resources. Second, we measured sparse overlap of facts expressed with each dimension across sources, which motivates future work on graph integration through (automated) node resolution. Third, comparing the dimension-based clustering to language model-based unsupervised edge clustering resulted in low overall agreement, though in some cases, the unsupervised clusters were dominated by one or two dimensions. This showed that some of the dimensions represent a stronger signal for language modeling than others. Fourth, we measured the impact of each dimension on a downstream question answering reasoning task, by adapting a state-of-the-art method of pretraining language models with knowledge graphs. Here, we observed that the impact differs greatly by dimension, depending largely on the alignment between the task and the knowledge dimension, as well as on the novelty of knowledge captured by a dimension. While this is in accordance with the findings of the original method~\cite{ma2020knowledgedriven}, the dimension-driven experiments of this paper enabled this hypothesis to be investigated much more precisely, revealing the direct impact of each knowledge dimension rather than entire knowledge sources.

Our experiments inspired a four-step roadmap towards creation and utilization of a comprehensive dimension-centered resource. (1) Node resolution methods should be introduced and applied to unify the resources further. (2) Such an integration would allow us to better understand and improve the coverage/gaps and boundaries of these sources. (3) A large-scale, public semantic graph of commonsense knowledge may inspire novel neuro-symbolic methods, potentially allowing for better generalization and explainability. (4) The impact of a dimension is an indicator of the coverage of that dimension in current evaluation benchmarks; under-represented dimensions are evaluation gaps that may need to be filled by introducing new benchmarks. And, vice-versa, additional knowledge dimensions might be hidden behind the generic \textit{relational-other} dimension.

\section*{Acknowledgement}
Four of the authors (Filip Ilievski, Bin Zhang, Deborah McGuinness, and Pedro Szekely) are sponsored by the DARPA MCS program under Contract No. N660011924033 with the United States Office Of Naval Research.

%\printcredits

\bibliographystyle{cas-model2-names}
\bibliography{refs}

\end{document}